\documentclass{article}
\PassOptionsToPackage{numbers, compress}{natbib}

\usepackage[final]{neurips_2022_ml4ad}

\usepackage[utf8]{inputenc} 
\usepackage[T1]{fontenc}    
\usepackage{hyperref}       
\usepackage{url}            
\usepackage{booktabs}       
\usepackage{amsfonts}       
\usepackage{nicefrac}       
\usepackage{microtype}      
\usepackage{xcolor}         
\usepackage{mathtools}
\usepackage{listings}
\usepackage{subfigure}
\usepackage{comment}
\usepackage{amssymb}
\usepackage{multirow}
\usepackage{tikz}
\usepackage{url}            

\usepackage{doi}

\usepackage[english]{babel}
\usepackage{changepage}
\usepackage{caption}
\usepackage{xspace}

\newcommand{\AB}[1]{\textcolor{black}{#1}}

\newcommand{\parag}[1]{\smallskip\noindent\textbf{#1}~~}

\newcommand{\real}{\mathbb{R}}

\newcommand{\vy}{\mathbf{y}}
\newcommand{\vx}{\mathbf{x}}

\definecolor{ForestGreen}{RGB}{34,139,34}
\definecolor{codecomment}{rgb}{0.6,0.6,0.53}
\definecolor{codekeyword}{rgb}{0.65,0.1,0.1}
\definecolor{codeblue}{rgb}{0.25,0.5,0.5}
\definecolor{codegreen}{rgb}{0,0.6,0}
\definecolor{codegray}{rgb}{0.5,0.5,0.5}
\definecolor{codepurple}{rgb}{0.58,0,0.82}
\definecolor{backcolour}{rgb}{0.95,0.95,0.92}

\lstdefinestyle{mystyle}{
    backgroundcolor=\color{white},   
    commentstyle=\fontsize{7.2pt}{7.2pt}\color{codeblue},
    keywordstyle=\fontsize{7.2pt}{7.2pt}\color{codekeyword},
    numberstyle=\tiny\color{codegray},
    stringstyle=\color{codepurple},
    basicstyle=\fontsize{7.2pt}{7.2pt}\ttfamily\selectfont,
    columns=fullflexible,
    breakatwhitespace=false,         
    breaklines=true,                 
    captionpos=b,                    
    keepspaces=true,                 
    numbers=left,                    
    numbersep=5pt,                  
    showspaces=false,                
    showstringspaces=false,
    showtabs=false,                  
    tabsize=2
}

\title{Improving Predictive Performance and Calibration by Weight Fusion in Semantic Segmentation}

\author{%
  Timo Sämann \\
  Valeo Schalter und Sensoren GmbH \\
  Germany \\
  \texttt{timo.saemann@valeo.com} \\
  \And
  Ahmed Mostafa Hammam \\
  Stellantis, Opel Automobile GmbH \\
  Germany \\
  ahmedmostafa.hammam@external.stellantis.com \\
  \And
  Andrei Bursuc \\
  valeo.ai \\ 
  France \\
  andrei.bursuc@valeo.com \\
  \And
  Christoph Stiller \\
  Institute of Measurement and Control Systems, Karlsruhe Institute of Technology \\
  Germany \\
  stiller@kit.edu \\
  \And
  Horst-Michael Groß \\
  Ilmenau University of Technology, Neuroinformatics and Cognitive Robotics Lab \\ 
  Germany \\
  Horst-Michael.Gross@tu-ilmenau.de \\
}

\begin{document}

\maketitle

\begin{abstract}
Averaging predictions of a deep ensemble of networks is a popular and effective method to improve predictive performance and calibration in various benchmarks and Kaggle competitions.
However, the runtime and training cost of deep ensembles grow linearly with the 
size of the ensemble, making them unsuitable for many applications.
Averaging ensemble weights instead of predictions circumvents this disadvantage during inference and is typically applied to intermediate checkpoints of a model to reduce training cost. Albeit effective, only few works have improved the understanding and the performance of weight averaging.
Here, we revisit this approach and show that a simple weight fusion (WF) strategy can lead to a significantly improved predictive performance and calibration. 
We describe what prerequisites the weights must meet in terms of weight space, functional space and loss. Furthermore, we present a new test method (called oracle test) to measure the functional space between weights. 
We demonstrate the versatility of our WF strategy across state of the art segmentation CNNs and Transformers as well as real world datasets such as BDD100K and Cityscapes. We compare WF with similar approaches and show our superiority for in- and out-of-distribution data in terms of predictive performance and calibration.
\end{abstract}

\section{Introduction}
Real-world applications for which AI systems are in the safety-critical path, i.e., represent the basis for the decision-making process, impose the highest demands on the 
AI modules composing these systems. This refers to generalization capability, robustness, calibration, and for practical reasons, computational efficiency. Deep ensembles can demonstrably contribute to an improvement in the mentioned points~\cite{ovadia2019can, gustafsson2020evaluating} except for efficiency. For a variety of applications, e.g., autonomous driving and driving assistance systems, the efficiency consideration is not insignificant. 
Despite a steadily increasing performance of information processing capabilities by both hardware and software, the required computing power and runtime is a tremendous challenge for a large number of applications. 
Fusion of weights instead of predictions or logits~\cite{lakshminarayanan2017simple,gal2016dropout} ultimately leads to 
\AB{a single set of weight only, thus effectively addressing the limited computational budget requirement.} 
\AB{The runtime of fused weights is exactly the same as for a single network, with the additional benefit of encapsulating diverse information from multiple networks.}

Fort et al.~\cite{fort2019deep} show that although the same Deep Neural Network (DNN)
architecture was trained with the same data and hyperparameters (only the initialization was different), the weak spots of the resulting weights were different. The basic idea of averaging weights is to compensate the weak spots of the individual weights. 
Averaging weights have been shot to lead to wider optima and better generalization~\cite{izmailov2018averaging}.
So far, few works have proposed a better understanding and improvements of this strategy. In addition, weight averaging has been mostly applied 
to classification tasks and the breakthrough to real world tasks such as semantic segmentation has been largely missing. One reason for the missing breakthrough may be that too little is known about the prerequisites of the weights to be fused and how these prerequisites could be fulfilled.

Here we focus on the real world task of semantic segmentation and shed light on the prerequisites of the weights to be fused. These preconditions include measurements in the weights space and function space between the weights. We measure the weight space using the cosine similarity which should be as high as possible for the weight fusion. At the same time, the weights in function space should be as far away as possible, since the distance correlates with the diversity of the weak spots. The higher the diversity, the higher the potential for weight fusion. 
For the measurement of the function space, we introduce the so-called oracle testing, which is novel to the best of the authors' knowledge. 
We give insights and guidance to find the optimal compromise between weight space and function space and describe two approaches to generating the weights to be fused. 
We apply weight fusion to three SOTA architectures: DeepLabV3+~\cite{chen2018encoder}, HRNet~\cite{wang2020deep}, and Segmenter~\cite{strudel2021segmenter}. We use the BDD100K~\cite{yu2018bdd100k}, Cityscapes~\cite{cordts2016cityscapes} and ACDC~\cite{sakaridis2021acdc} dataset for our experiments. 

Our main contribution is to demonstrate a simple and effective method to fuse two or more 
sets of DNN weights into a single one.
\textbf{(i)} Our weight fusion method improves predictive performance and calibration without impacting runtime cost. 
\textbf{(ii)} In extensive studies, we derive important properties of 
DNN weights towards
weight fusion. \textbf{(iii)} We introduce a new testing method that can measure the functional space between weights, called oracle testing. 
\textbf{(iv)} We 
conduct extensive experiments on multiple SoTA segmentation architectures (CNNs and Transformers) on multiple real world datasets. \textbf{(v)} We show the superiority of our approach in a comparison with Stochastic Weight Averaging (SWA)~\cite{izmailov2018averaging} and deep ensembles for in-distribution as well as for out-of-distribution data (ACDC)~\cite{sakaridis2021acdc}.

\section{Weight Fusion}
Our approach of weight fusion is to generate at least two weights of the same DNN and to fuse them with a weighted averaging into a single network. The goal of the weight fusion is to improve the performance as well as the calibration of the DNN, while not impacting the runtime performance of the DNN.

Formally, we define a training dataset $\mathcal{D} = \{ (\vx_i, \vy_i) \}_{i=1}^{n}$ with $n$ samples and labels, where $\vx_i \in \real^{h\times w \times c}$ are images with spatial resolution $h \times w$ and $c$ color channels, and $\vy_i \in \real^{h \times w \times 1}$ are semantic segmentation labels (one label for each pixel in $\vx_i$). We consider a neural network $f_{\theta}(\cdot)$ with trainable parameters $\theta$, that processes the input image $\vx_i$ and outputs a semantic map prediction $\hat{\vy}_i$. Different random initializations of the parameters $\theta$ lead to different sets of weights after training. We denote with $\theta^{(\tau)}$ the parameters of a specific training process $\tau$ and with $f_{\theta^{(\tau)}}(\cdot)$ the corresponding network.

The weight fusion represents a weighted averaging of the weight files (checkpoints) computed from two different training runs. Formally we define the element-wise fusion of two sets of parameters as follows: $\bar{\theta} =  \alpha \theta^{(1)} + \beta \theta^{(2)}$,
where $\theta^{(1)}$ and $\theta^{(2)}$ are two sets of parameters for $f$, $\alpha$ and $\beta$ are fusion parameters with $\beta=1-\alpha$, and $\bar{\theta}$ is the set of fused weights.




\subsection{Weight Properties for Improved Fusion}
We hypothesize that weights $\theta^{(\tau)}$ must satisfy three properties to achieve optimal performance and calibration improvements: (i) \textbf{high cosine similarity} to each other; (ii) a \textbf{high value in oracle testing}; (iii) a \textbf{low loss on the validation data}.
In the following we 
\AB{detail} these properties.

\parag{(i) Cosine similarity.} Cosine similarity is a measure of the similarity of two vectors. The cosine of the angle between two vectors $\theta^{(1)}$ and $\theta^{(2)}$ (we flatten and concat the DNN parameters into a single vector) is determined by: 
\begin{math}
\cos(\measuredangle) = \frac{\mathbf{\theta^{(1)}} \cdot \theta^{(2)}}{||\theta^{(1)}|| \cdot ||\theta^{(2)}||} 
\end{math}.
An angle of 0 degrees corresponds to 1, i.e., the vectors are parallel, while an angle of 90 degrees results in 
0, i.e., the vectors are perpendicular to each other.
Cosine similarity can be considered as a measurement in the weight space. DNNs with different initializations can reach highly different local optima after training. Two networks can have very similar predictions but low similarity, e.g., convolutional filters from the two networks are permuted. Since the weight fusion is done element-wise it is essential for the two networks to not be far away from each other as the final weights $\bar{\theta}$ risk being useless.
We add the pytorch code snippets for weight fusion and cosine similarity in Appendix~\ref{pytorch_code_snippet}.

\parag{(ii) Oracle testing.} In Oracle testing, an oracle merges all true positives of the DNN predictions into a single output/prediction based on the ground truth. Then, the corresponding metric is calculated for the merged prediction (e.g., mIoU). If the DNN predictions are very similar in their errors, the result of the metric will be similar to the single predictions. In other words, the DNNs are close together in the functional space. If the DNN predictions are diverse in terms of true positives, the oracle test yields a high value.
This means that the DNNs are far away in the function space.
The quality of an ensemble is a major advantage of 
DE~\cite{lakshminarayanan2017simple} over other approaches~\cite{fort2019deep}. Higher diversity means that the models are not stuck in the same local optimum and do not have the same weak spots. Quantitatively, diversity is usually computed as dissimilarity between predictions from pairs of networks~\cite{aksela2003comparison,fort2019deep,rame2021dice}. Oracle testing takes the ground truth into account and can be considered as a measurement in the function space. The pytorch code snippet is listed \AB{below} in~\ref{lst:oracle}.

\begin{lstlisting}[basicstyle=\small, language=Python, label={lst:oracle}, caption={Iterate over the number of weights to be used to calculate the oracle testing. \textit{Outputs} is a list containing the output of the models.}, captionpos=b]
for num in range(number_weights):
    mask[num] = [outputs[num] == ground_truth] 
    outputs[0][mask[num]] = outputs[num][mask[num]]
\end{lstlisting}

\parag{(iii) Validation loss.}
Empirically, we can show that the validation loss of the weights to be fused is a relevant factor and should be close to its local minimum.

\subsection{Generation of DNN Weights}
To generate the weights, we distinguish the following two options, which we examine in more detail in sec.~\ref{sec:experiments}.

1) The weights are generated by training from 
\AB{``scratch''} (see~\S\ref{sec:from_sratch}). Here it is necessary to keep the hyperparameters as well as the training data the same for each training run. To achieve a high oracle test result, 
\AB{we show} in our experiments with DeeplabV3+, HRNet, and the Segmenter that changing the initialization seed is sufficient. This change does reduce cosine similarity, but only slightly. With this approach we achieve a good compromise between cosine similarity and oracle testing for the mentioned architectures.

2) The weights are derived from a finetuning where 
\AB{an already} trained weight is used as starting weight (see~\S\ref{finetuning}). The same data is used for the finetuning as for the \AB{prior} training.
The finetuning is performed with a cosine annealing learning rate schedule and can range from one to several cycles. The cosine annealing learning rate schedule is defined as:  
$\text{lr} = \frac{1}{2} \cdot \sum_{T=1}^{N} \text{start\_lr} \cdot \left(1 + \frac{T}{N \cdot \pi}\right) $.
$T$ is the current 
iteration and $N$ is the total amount of iterations, i.e., the multiplication of the number of epochs by the number of iterations per epoch.
The length of the cycles can range from one to several epochs. The height of the start-learning rate for the cycles is crucial for the optimal property of the weights for the fusion.

\parag{Further procedure.} In 1) and 2) we use the presented properties (cosine similarity, oracle testing, validation loss) of the weights to determine the suitable weights for the fusion. Then, we determine the fusion parameter 
\AB{$\alpha$} using grid search in 0.05 steps on the validation data. Please note that training the 
\AB{$\alpha$} parameter would not lead to any noteworthy improvement for the following two reasons: (i) Finer resolution of the step size did not achieve any significant improvements $ \leq +0.1 $ mIoU). (ii) There is little reduction in computational effort when comparing the forward passes required for the grid search with the forward plus backward passes required in a training process over multiple epochs.
Once several fused weights have been determined, one selects the fused weight with the best predictive performance on the validation data as the final fused weight. 
This procedure is transferable to new DNNs, datasets or tasks.

\section{Experiments}\label{sec:experiments}
The generation of the weights to be fused can be done in two ways. (i) By training from 
\AB{``scratch''} and (ii) finetuning. Therefore, we divide the experiments into two subsections. 
\S\ref{sec:from_sratch} describes weight fusion based on weights obtained through multiple trainings from 
scratch. We show that our weight fusion approach generalizes to multiple DNN architectures as well as datasets. We demonstrate statistically significant results for three state-of-the-art architectures for semantic segmentation DeeplabV3+~\cite{chen2018encoder}, HRNet~\cite{wang2020deep} and Segmenter~\cite{strudel2021segmenter}, based on real world datasets such as BDD100K and Cityscapes. 
We consider the fusion properties for performing weight fusion, but not focusing on analysing them in this subsection.

In 
\S\ref{finetuning}, we use a final trained weight as a starting point for finetuning. We demonstrate detailed analysis on cosine similarity, oracle testing and validation loss in the context of weight fusion. 
For this and subsequent experiments, we limit ourselves to using DeeplabV3+ and BDD100K due to 
\AB{limited compute budget.}

To avoid negative influence of label errors in the ground truth data of the BDD100K dataset, we adapted it manually. Furthermore, we split the available 7k images into 5k training data, 1k validation data and 1k test data. The process for adaptation and splitting as well as a description of Cityscapes~\cite{cordts2016cityscapes} and ACDC~\cite{sakaridis2021acdc} is described in more detail in Appendix~\ref{adapt_bdd}.
For reproducibility of the experiments, we list all the hyperparameters used for the trainings in Appendix~\ref{Used_hyperparameters}.

\subsection{Fusion of Weights Trained from Scratch}
\label{sec:from_sratch}
For the generation of the weights 
we run several trainings from scratch. 
We change only the initialization by specifying different random seeds, i.e., the architecture, the hyperparameters and the training data remain exactly the same.
These constraints lead to 
weight sets with a good compromise between oracle testing and cosine similarity, which is needed for weight fusion. To ensure a certain statistical significance, we performed four 
training runs each with DeepLabV3+, HRNet and Segmenter on the BDD100K and Cityscapes datasets - a total of 20 
runs.\footnote{Due to convergence difficulties with the Segmenter on the BDD100K data, we limit the evaluation of weight fusion to Cityscapes.} This serves to reduce statistical fluctuations due to different sources of variation 
\cite{dehghani2021benchmark} and 
allowing us to reliably attribute the performance improvement to weight fusion only.
The resulting 
4 weights per DNN architecture and dataset allow for 6 different possible combinations when fusing 2 weights each according to:  \begin{math} \frac{n!}{r!(n-r)! }\end{math} with n = 4 and r = 2.

\begin{figure}[t]
    \centering
    \subfigure[]{\includegraphics[width=0.45\textwidth]{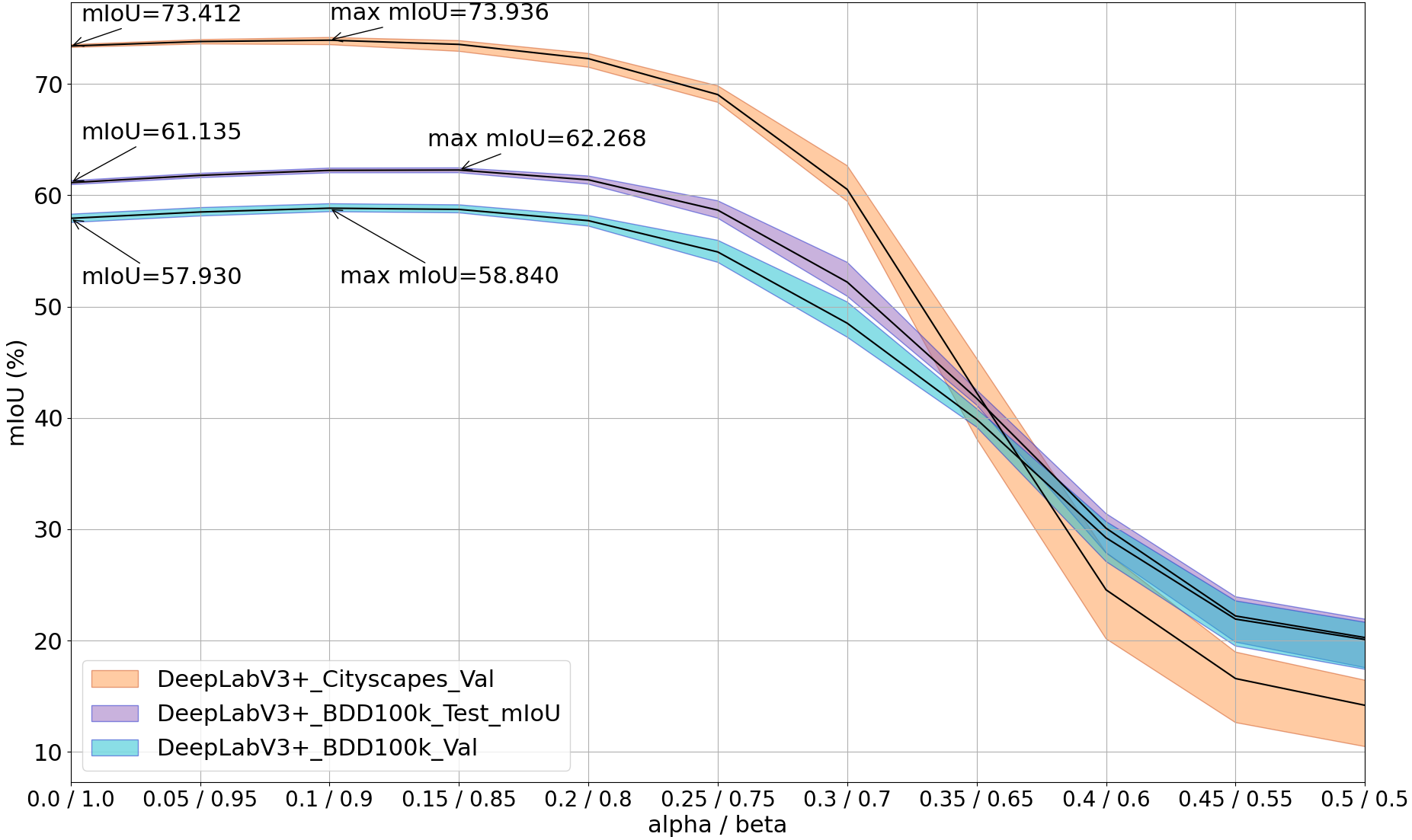}} 
    \subfigure[]{\includegraphics[width=0.45\textwidth]{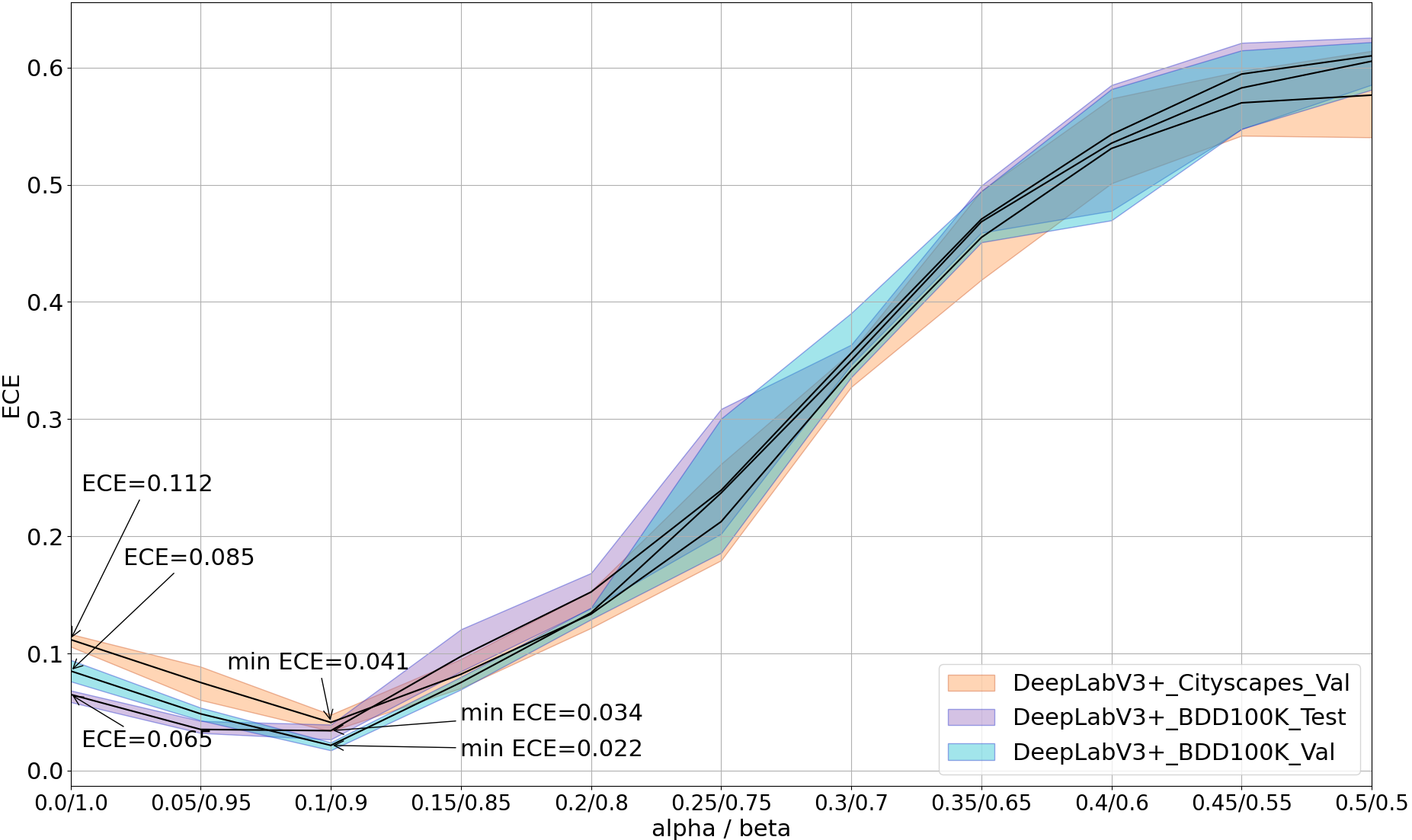}} 
    \quad
    \caption{DeepLabV3+ performance and calibration for 6 fused weights averaged for 
    \AB{$\alpha$} from 0 to 1 in 0.05 steps. Please note that 
    \AB{$\alpha /\beta$} = 0/1 corresponds to the single weights. \textit{(a)} mIoU \textit{(b)} ECE.}
    \label{fig:performance_from_sratch}
\end{figure}

Fig.~\ref{fig:performance_from_sratch} illustrates the behavior of the DeepLabV3+ architecture during weight fusion in terms of predictive performance and calibration.
Fig.~\ref{fig:performance_from_sratch}(a) shows the mIoU for DeepLabV3+ on the validation data from BDD100K and Cityscapes and on the test data from BDD100K. The mIoU values are obtained by averaging the mIoU values over the 
6 fused weights and are represented by a black line. The colored areas show the minimum and maximum mIoU values. The x-axis describes 
\AB{$\alpha$ and $\beta$}, the scalar values used as fusion parameter. The mIoU values are mirrored at the point 
\AB{$\alpha$} = 0.5, so that the order of the weights can be neglected. For this reason, the x-axis that extends to 0.5 is complete.
It is illustrated that with the fusion parameters 
\AB{$\alpha /\beta$} = 0.1/0.9 and 0.15/0.85, the maximum mIoU value is obtained over the 6 averaged mIoU values (see arrows in Fig.~\ref{fig:performance_from_sratch}(a)).
Table~\ref{tab:miou_Deeplab_BDD_City} shows the classes mIoU for these fusion parameters. It can be seen that the classes with a low pixel density such as 
\AB{\texttt{traffic light}, \texttt{traffic sign}, \texttt{bicycle} and \texttt{person}} benefit the most from weight fusion.

\begin{table}[t]
\caption{Per class mIoU in \% with the fusion parameters 
\AB{$\alpha /\beta$} = 0.1/0.9 for BDD100K and 0.15/0.85 for Cityscapes. The number 
between brackets shows the improvement (green) or deterioration (red) compared to the single weights. Especially the classes with a small pixel density have improved the most. Please note that the class 
\AB{\texttt{train}}
was not learned at all probably due to the limited images containing 
\AB{\texttt{train}} in the BDD100K training data.}
\hrule
\label{tab:miou_Deeplab_BDD_City}
\centering
\resizebox{\textwidth}{!}{%
\begin{tabular}{c||c|c|c|c|c|c|c|c|c|c|c|c|c|c|c|c|c|c|c||c}
                            & Road    & Sidewalk & Building & Wall    & Fence   & Pole    & Traffic Light & Traffic Sign & Vegetation & Terrain & Sky     & Person  & Rider   & Car     & Truck   & Bus     & Train   & Motorcycle & Bicycle & Overall \\ \hline
\multirow{2}{*}{BDD100K}    & 94.71   & 65.40    & 85.77    & 29.29   & 51.14   & 52.91   & 58.33         & 56.29        & 86.49      & 51.17   & 95.32   & 65.73   & 46.36   & 90.70                        & 58.88   & 81.41   & 0.00    & 55.87      & 57.30   & 62.27   \\
                            & \color{ForestGreen}(+0.14) & \color{ForestGreen}(+0.51)  & \color{ForestGreen}(+0.25)   & \color{ForestGreen}(+0.02) & \color{ForestGreen}(+0.14) & \color{ForestGreen}(+2.86) & \color{ForestGreen}(+4.67)       & \color{ForestGreen}(+3.46)      & \color{ForestGreen}(+0.12)    & \color{ForestGreen}(+0.56) & \color{ForestGreen}(+0.17) & \color{ForestGreen}(+2.10) & \color{red}(-1.33) & \multicolumn{1}{c|}{\color{ForestGreen}(+0.39)} & \color{ForestGreen}(+1.73) & \color{ForestGreen}(+2.30) & (0.00)  & \color{ForestGreen}(+1.42)    & \color{ForestGreen}(+3.67) & \color{ForestGreen}(+1.14) \\ \hline
\multirow{2}{*}{Cityscapes} & 97.73   & 82.92    & 91.39    & 52.72   & 56.77   & 60.09   & 67.03         & 75.64        & 91.56      & 62.95   & 94.06   & 79.41   & 60.62   & \multicolumn{1}{c|}{93.48}   & 64.25   & 80.67   & 61.55   & 58.86      & 74.59   & 73.94   \\
                            & \color{ForestGreen}(+0.03) & \color{red}(-0.01)  & \color{ForestGreen}(+0.25)  & \color{ForestGreen}(+0.44) & \color{red}(-0.23) & \color{ForestGreen}(+0.78) & \color{ForestGreen}(+2.93)       & \color{ForestGreen}(+1.67)      & \color{ForestGreen}(+0.17)    & \color{ForestGreen}(+0.14) & \color{ForestGreen}(+0.60) & \color{ForestGreen}(+1.25) & \color{ForestGreen}(+1.00) & \multicolumn{1}{c|}{\color{ForestGreen}(+0.01)} & \color{red}(-0.35) & \color{ForestGreen}(+0.25) & \color{ForestGreen}(+0.99) & \color{ForestGreen}(+0.20)    & \color{ForestGreen}(+1.36) & \color{ForestGreen}(+0.53)
\end{tabular}}
\end{table}

\begin{figure}[t]%
\centering
    \subfigure[]{\includegraphics[width=0.45\textwidth]{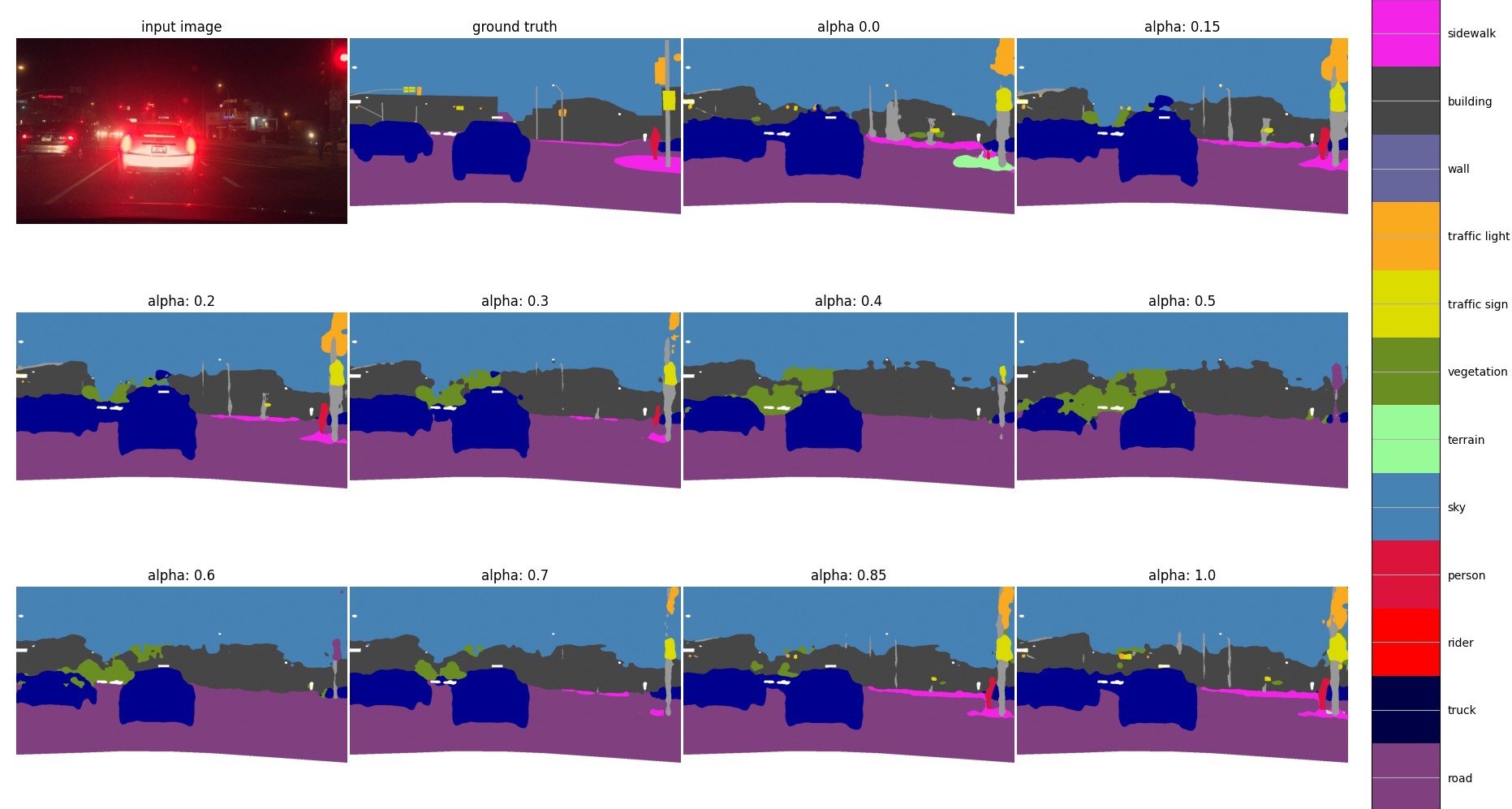}}%
    \vspace{0.01pt}
    \subfigure[]{\includegraphics[width=0.45\textwidth]{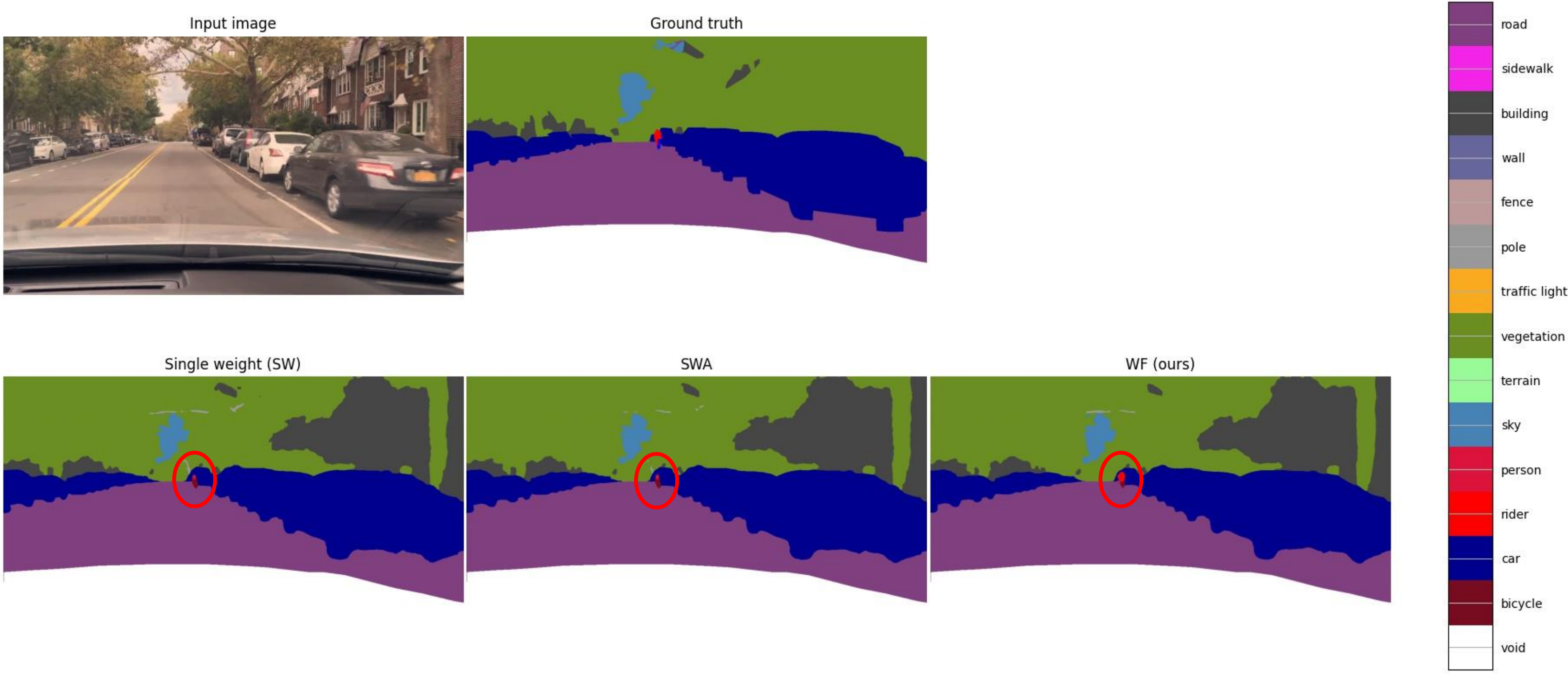}}

\caption{(a) Illustration of semantic segmentation for different alpha values using DeepLabV3+ and an example image from BDD100K. 
    \AB{$\alpha$} = 0.0 and 1.0 show the predictions of the single weights. 
    \AB{$\alpha$} = 0.15 shows the prediction of the fused weight where the pedestrian and the sidewalk are better segmented. Please note that the color white represents 
    \AB{\texttt{void}}. (b) Visual comparison of the SWA, weight fusion (WF) and single weight (SW) methods. WF is able to segment the rider correctly (see ellipse).}
\label{fig:deeplab_alpha_all}
\end{figure}

How this behavior looks in semantic segmentation masks can be seen in Fig.~\ref{fig:deeplab_alpha_all}(a). The images with 
\AB{$\alpha$} = 0 and 
\AB{$\alpha$} = 1 correspond to the single predictions of weights 0 and 1, respectively. As alpha approaches 0.5, fewer predictions are seen for the small classes such as 
\AB{\texttt{traffic light}, \texttt{person},} etc., which explains the increased precision and reduced recall.
Looking at the 
\AB{\texttt{sidewalk}} in the image with 
\AB{$\alpha$} = 0.15, we notice that it is most similar to ground truth. This shows that the fused weight is not simply an alignment of the two weights 0 and 1, but represents a completely new function. In other words, the weight fusion results in a new point in function space. More figures confirming this behavior can be found in Appendix~\ref{Further_results}.

Further analysis was done on the weight fusion by evaluating the calibration of the fused DNNs based on the expected calibration error (ECE)~\cite{guo2017calibration, neumann2018relaxed}, see Fig.~\ref{fig:performance_from_sratch}(b). Explanation of the ECE are in Appendix~\ref{calibration_ece} available. It can be seen that the lowest ECE is found at 
\AB{$\alpha$} = 0.1 where the ECE reaches the lowest value for all three data splits, which means that a significantly better calibration is achieved matching its counterpart maximum mIoU in Fig.~\ref{fig:performance_from_sratch}(a). This indicates that the fusion does not only improve the mIoU but also improves the calibration of the network. Further evaluations regarding precision and recall as well as Kullback-Leibler (KL) divergence can be found in Appendix~\ref{precision_recall_kl}.

Average results for DeepLabV3+, HRNet, and Segmenter are shown Table~\ref{tab:results_for_three_architectures}. The numbers in parentheses indicate the improvement (green) and deterioration (red) from the single weight. Please note that the numbers given are averages over 4 single weights and the resulting 6 fused weights. Furthermore, for the sake of clarity, the same 
\AB{$\alpha$} parameter was used for all 6 fused weights, i.e., there was no individual selection of 
\AB{$\alpha$}, which would have led to even better results. For the results on BDD100K, the 
\AB{$\alpha$} value that performed best on the validation data was used.
Improvements in performance (mIoU, precision, recall) and calibration (ECE, KL) can be seen across the board. This demonstrates that the weight fusion is also transferable to different CNNs and also Transformers.

\setlength{\tabcolsep}{8pt}
\begin{table}[t]
\caption{Results for three architectures.}
\label{tab:results_for_three_architectures}
\centering
\resizebox{\textwidth}{!}{%
\begin{tabular}{ccc|cc|c}
\toprule
   & \multicolumn{2}{c}{DeepLabV3+} & \multicolumn{2}{c}{HRNet}  & \multicolumn{1}{c}{Segmenter} \\
    \cmidrule(r){2-3}
    \cmidrule(r){4-5}
    \cmidrule(r){5-6}

 & \multicolumn{1}{|c}{BDD100K}  & Cityscapes & BDD100K  & Cityscapes & Cityscapes  \\ 
 \cmidrule(r){2-3}
    \cmidrule(r){4-5}
    \cmidrule(r){5-6}
Metrics & \multicolumn{1}{|c}{alpha=0.1} & alpha=0.1 & alpha=0.45 & alpha=0.1 & alpha=0.05 \\ \hline
\multicolumn{1}{r|}{Mean IoU (\%) ($\uparrow$)} & 62.24 \color{ForestGreen}(+1.10) & 73.94 \color{ForestGreen}(+0.53) 
& 55.57\color{ForestGreen}(+0.79) & 69.65\color{ForestGreen}(+0.23) & 70.88 \color{ForestGreen}(+0.34)\\
\multicolumn{1}{r|}{Precision (\%) ($\uparrow$)}   & 75.28 \color{ForestGreen}(+2.92)  & 86.47 \color{ForestGreen}(+2.19) 
& 72.05\color{ForestGreen}(+0.46) & 83.68\color{ForestGreen}(+1.20) & 84.14\color{ForestGreen}(+0.30)\\
\multicolumn{1}{r|}{Recall (\%) ($\uparrow$)}  & 72.69 \color{red}(-1.34)  & 82.66 \color{red}(-1.40) 
& 65.31\color{ForestGreen}(+1.56) & 79.86\color{ForestGreen}(+0.63) & 80.24\color{ForestGreen}(+0.22)\\
\multicolumn{1}{r|}{ECE ($\downarrow$)} & 0.034 \color{ForestGreen}(-0.031)  & 0.041 \color{ForestGreen}(-0.071)
& 0.111\color{ForestGreen}(-0.002) & 0.132\color{ForestGreen}(-0.007) & 0.030\color{red}(-0.001)  \\
\multicolumn{1}{r|}{KL ($\uparrow$)}          & 1.919 \color{ForestGreen}(+0.312)  & 1.826 \color{ForestGreen}(+0.297) 
& 0.858\color{red}(-0.077) & 0.952\color{ForestGreen}(+0.07) & 2.445\color{ForestGreen}(+0.016)
\end{tabular}}
\end{table}

\subsection{Fusion of Finetuned Weights}
\label{finetuning}
Besides training from scratch~(see \S\ref{sec:from_sratch}), finetuning is a way to generate the weights for the WF. Finetuning uses the same data as training from scratch. A fully trained weight is used for initialization which we consider as starting weight. In our experiments the finetuning is performed with learning rates between 0.002 and 0.02 in 0.002 steps and covers 10 epochs. The cosine annealing learning rate schedule was used so that the 10 epochs correspond to one cycle (see Fig.~\ref{fig:learning_rate_schedule_color_and_black}(a)). 
For the fusion we use the starting weight (at epoch 50) and one of the individual checkpoints of the respective epochs 1 to 10. For this experiment we use the DeepLabV3+ and the BDD100K validation data set. 
Fig.~\ref{fig:finetuning_experiments}(a) and (b) show the cosine similarity and the mIoU of the oracle testing, respectively, for the starting weight and the weight of the 10th epoch.
Fig.~\ref{fig:finetuning_experiments}(c) shows the mIoU after fusion of the starting weight with the checkpoint of the 10th epoch along the fusion parameter 
\AB{$\alpha$} for the corresponding learning rates. 
The improvements of the mIoU compared to the starting weight, which is 58.61~\%, are marked with a diamond. From Fig.~\ref{fig:finetuning_experiments} we derive the following findings:
\begin{figure}[t]
    \centering
    \subfigure[]{\includegraphics[width=0.4\textwidth]{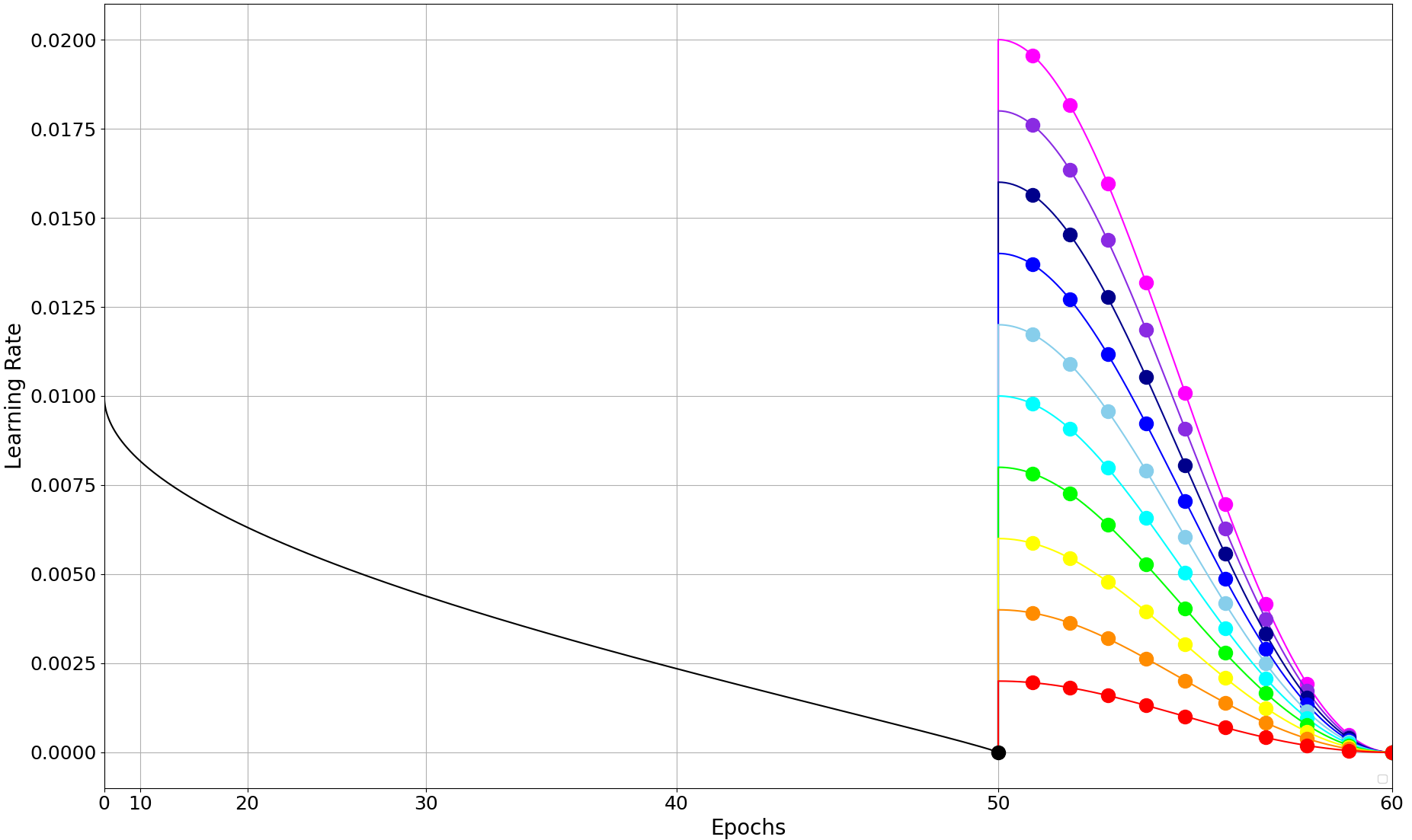}}
    \subfigure[]{\includegraphics[width=0.4\textwidth]{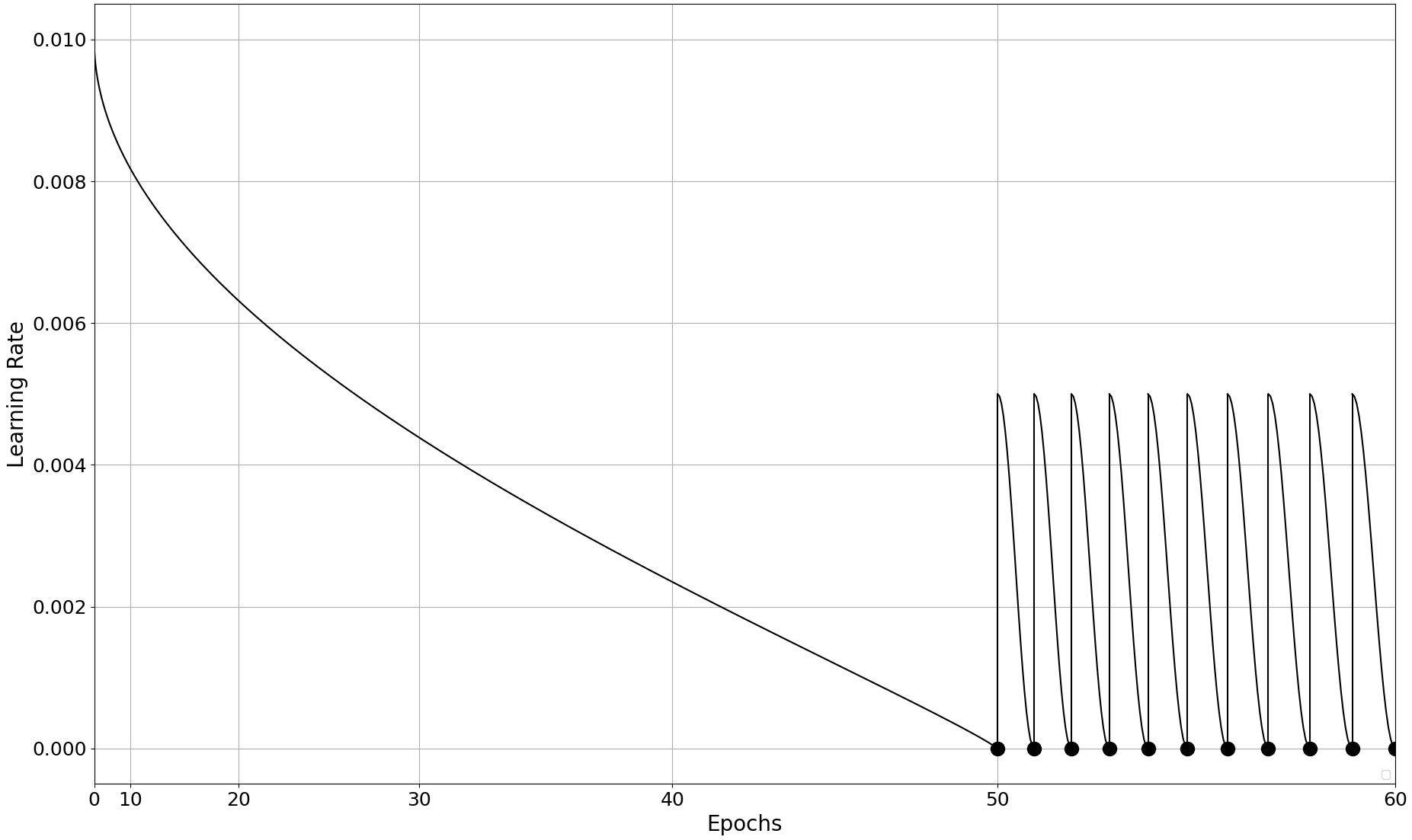}}
    \caption{Representation of the learning rate schedule. The blobs indicate the positions from which a checkpoint was stored for our experiments. (a) Used in \S\ref{finetuning}. (b) Used in \S\ref{sec:comparison_to_SWA}}
    \label{fig:learning_rate_schedule_color_and_black}
\end{figure}

\begin{figure}[h!]
    \centering
    \subfigure[]{\includegraphics[width=0.24\textwidth]{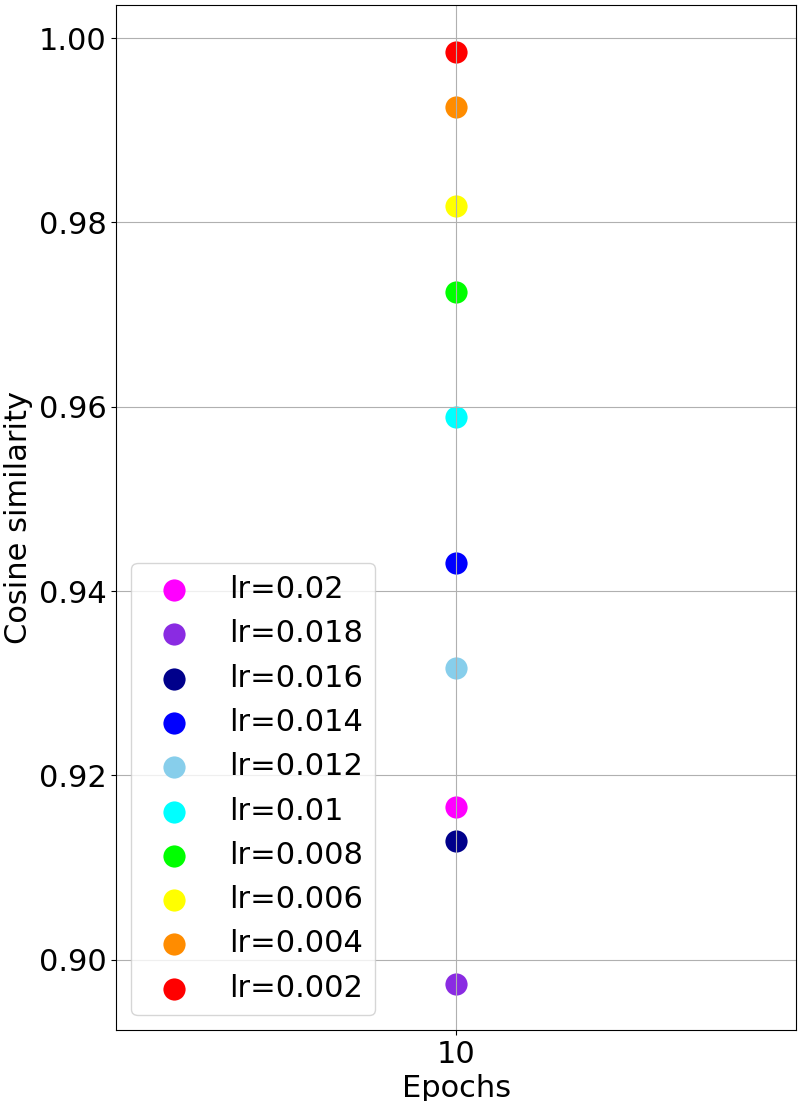}}
    \subfigure[]{\includegraphics[width=0.24\textwidth]{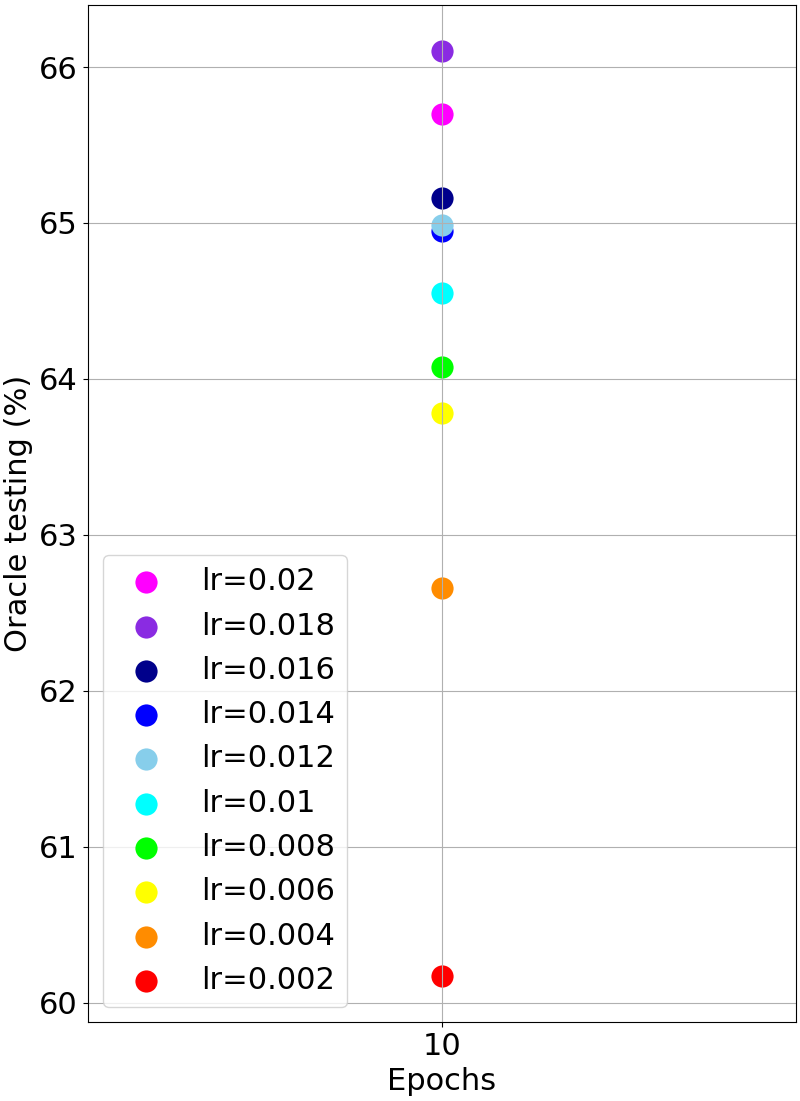}}
    \subfigure[]{\includegraphics[width=0.49\textwidth]{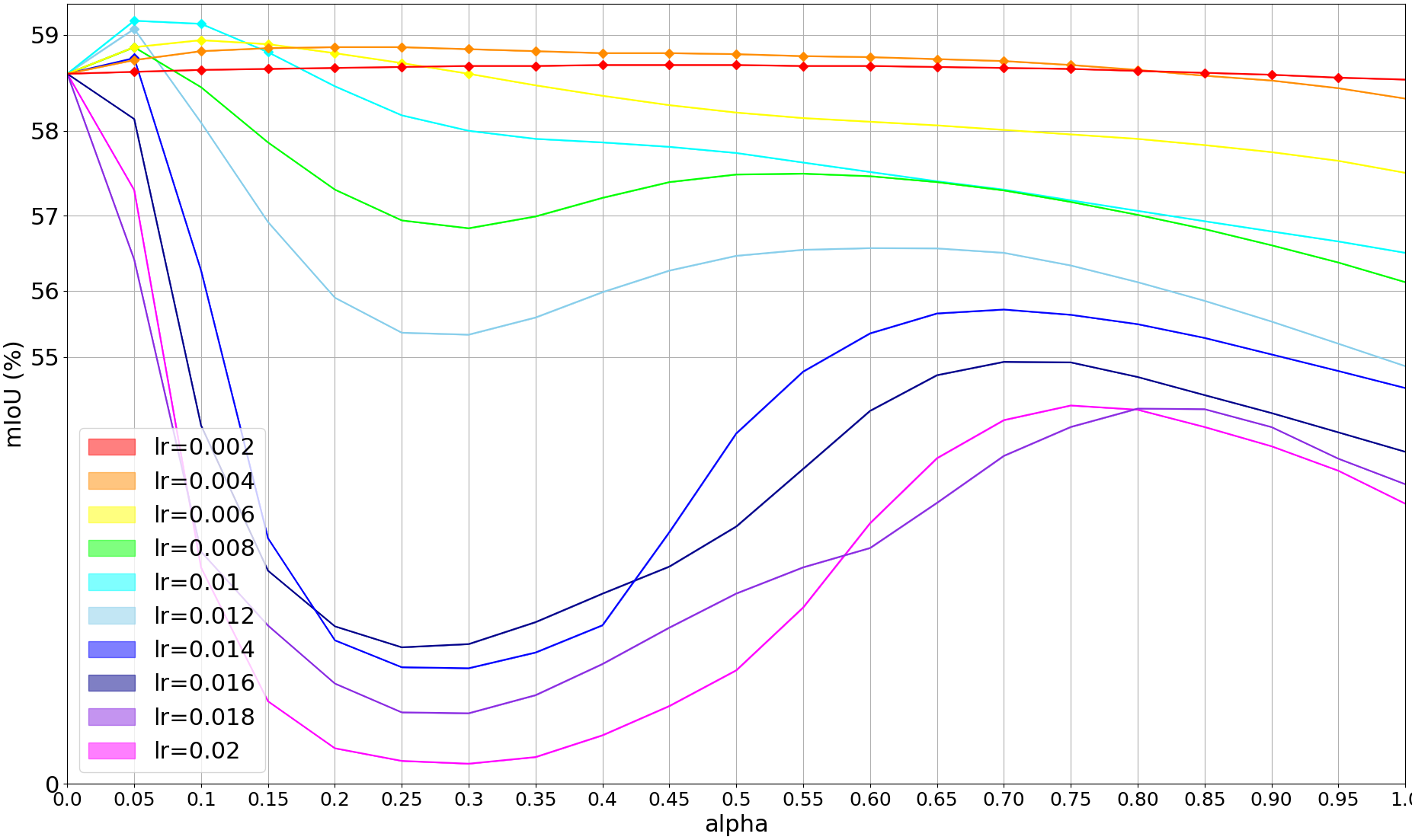}} 
    \caption{\textit{(a)} Cosine similarity for the 10th epoch weight compared to the starting weight for the corresponding learning rates. \textit{(b)} Oracle testing for the 10th epoch weight and the starting weight for the corresponding learning rates. \textit{(c)} mIoU after fusion of the starting weight with the checkpoint after the 10th epoch along the fusion parameter alpha for the corresponding learning rates.}
    \label{fig:finetuning_experiments}
\end{figure}
\vspace{-4mm}

\subsection*{Finding 1: Cosine similarity and oracle testing are correlated.}
The cosine similarity measures the distance of the weights in the weight space and the oracle testing in the functional space. The closer the weights are in weight space, the closer they are in functional space and vice versa. Using Fig.~\ref{fig:finetuning_experiments}(a) and (b) it can be seen that the lower the learning rate the closer they are together in weight space (cosine similarity approaches 1.0) and at the same time they are close together in functional space (oracle testing approaches the single weight mIoU of 58.61 \%).

\subsection*{Finding 2: Cosine similarity must not be too low.}
A low cosine similarity is equivalent to a long distance in weight space. For a successful fusion the weights must not be too far apart in weight space.
Based on the mIoU of the fused weights (see Fig.~\ref{fig:finetuning_experiments}(a) and (c)) it can be seen that a cosine similarity smaller than about 0.925 at no 
\AB{$\alpha$} value results in a higher mIoU than the starting value. This is the case for learning rates 0.02, 0.018 and 0.016. 
\subsection*{Finding 3: Oracle test and cosine similarity should both be as high as possible.}
Based on Finding 1, we know that these are negatively correlated with each other and thus a compromise must be found. The learning rate 0.01 seems to be a good compromise, since it leads to the highest fused mIoU in Fig.~\ref{fig:finetuning_experiments}(c).

Two further findings regarding the weighting parameter and the validation loss have been derived and are described in Appendix~\ref{further_finding}.
By and large the combination of a low validation loss with a high oracle testing and a high cosine similarity are criteria for a successful weight fusion. This tradeoff is not achievable for SWA because the weights are equal-weighted averaged, which in turn requires a very high cosine similarity between the weights (cf. Finding 4 in~\ref{further_finding}). WF can find a better compromise by exploiting the
\AB{$\alpha$} parameter. To apply this knowledge to new problems, we have summarized some practical advice in Appendix~\ref{Practical_Advice}.

\section{Results}\label{sec:comparison_to_SWA}
For the results section, we restrict ourselves to the DeepLabV3+ architecture for our WF approach as well as for the comparison methods. 
We train all methods on the BDD100K training dataset and evaluate them using the test dataset from BDD100K and ACDC. The ACDC dataset is used to evaluate the generalization ability of our WF in the presence of a distribution shift.
The ACDC dataset contains 19 classes as BDD100K and 4 test datasets of the domains fog, night, rain and snow. 
First, in  
\S\ref{sec:results_method_decription} we describe the methods used for the results and then in
\S\ref{sec:qualitative_quantitative_results} we present mainly the qualitative results in Table~\ref{tab:overall_results}.

\subsection{WF and Comparison Methods}\label{sec:results_method_decription}
For the comparison we consider the SWA method as well as deep ensembles. 
For 
\AB{SWA} we train the DeepLabV3+ for 50 epochs and then 
\AB{finetune} with cosine annealing for 10 cycles, where one cycle corresponds to one epoch. After 50 epochs and at the end of each cycle, we 
\AB{save} the weights for the SWA baseline. This results in a total of 11 weights.
\AB{We show the scheduling of the learning rate} in Fig.~\ref{fig:learning_rate_schedule_color_and_black}(b)\footnote{For the 10 cycles, we tested the learning rates 0.0025, 0.005, and 0.0075. With the learning rate 0.005 we have achieved the best result for SWA and therefore we will focus on this learning rate in the following.}. The blobs in this figure indicate the weights used for the SWA baseline. We refer to the single weight after 50 epochs as SW.
To construct the SWA baseline, we perform an equally weighted averaging over the 11 weights. 
\AB{We subsequently adapt that statistics of the BatchNorm layers by updating the running means and variances over the training data~\cite{garipov2018loss, izmailov2018averaging}.}
However, this adaptation had very little impact on performance and calibration in our evaluation.  

For the comparison of our weight fusion (WF) to SWA, we choose 2 out of the 11 weights. We take the weight after 50 epochs (SW) and one of the remaining 10 weights. 
We test all 10 combinations and determine 
\AB{$\alpha$} by 
\AB{using} grid search in 0.05 steps on the validation data. We list an analysis for cosine similarity, oracle testing and validation loss for this in Appendix~\ref{further_analysis_wf}. For the test data, the same 
\AB{$\alpha$} parameter is used that was determined on the validation data. 

\AB{We also compare against DE of varying sizes: two, three and four member ensembles denoted as DE\_2, DE\_3 and DE\_4, respectively.}
\AB{We use the 4 DeepLabV3+ models from \S~\ref{sec:from_sratch} trained from scratch with different random initializations.}
The fusion is performed at softmax level, i.e., 
\AB{we average the softmax predictions of the members of the ensemble.}

For our comparison approaches (WF\_2, WF\_3, WF\_4) to the deep ensembles, we use the same weights. To determine the fusion parameter 
\AB{$\alpha$} for WF\_2, we perform a grid search in 0.05 steps using the training dataset and fuse the two weights accordingly. 
For WF\_3 and WF\_4, we need one and two additional fusion parameters, respectively. Since the computational cost of grid search increases exponentially with each additional fusion parameter, we selected the fusion parameters by a random process. Fusion parameters were chosen randomly and the mIoU were determined on the training data. This process was performed 50 times for WF\_3 and WF\_4, which kept 2 NVIDIA TITAN RTX GPUs busy for 1 day. The fusion parameter that achieved the highest mIoU on the training data was used to create the final weight for WF\_3 and WF\_4. The tables with the corresponding numerical data for this process to better understand the selection of members for DE\_2 and DE\_3 can be found in Appendix~\ref{comparison_to_de}.
We also use the 4 single weights used by the DE\_4 and WF\_4 as baselines. We 
\AB{report} the mean of the mIoU, ECE and KL metrics for the predictions of the 4 single weights. We denote this by 
\AB{``Mean SW''.}

\subsection{Qualitative and Quantitative Results}
\label{sec:qualitative_quantitative_results}

After explaining the methods used, we present the qualitative results on BDD100K and ACDC in Table~\ref{tab:overall_results}. For 
SWA and WF we 
\AB{report} the difference in brackets (green or red) with respect to SW.
For 
DE\_2 to WF\_4 we 
\AB{show} the difference in brackets (green or red) with respect to Mean SW. The change of the reference method from SW to Mean SW seems to be a logical consequence, since the 4 weight files of Mean SW are partially represented in DE\_2, DE\_3, WF\_2 and WF\_3 and completely in DE\_4 and WF\_4. 
Our WF method outperforms the single weight as well as our main competitor SWA in mIoU, ECE and KL. Looking at the test data of ACDC, the difference is significant. For example, in the domain 
\AB{\texttt{snow}} 
\AB{WF boosts mIou with +4.28\%,} while SWA \AB{is worse by -0.34\% mIoU than the single weight.}
The comparison of the calibration metrics ECE and KL show that WF is slightly better calibrated than SWA. 

\AB{We conduct a visual comparison between the SW, SWA and WF predictions in Fig.~\ref{fig:deeplab_alpha_all}(b).
}
Our approach predicts the 
\AB{\texttt{rider}} class (see red ellipse) more accurately. Please note that in the ground truth image the 
\AB{\texttt{bike}} was incorrectly labeled as 
\AB{\texttt{motorcyle}}, which is \AB{however} correctly predicted by all approaches. 

The evaluation of the weight fusion methods (WF\_2 to WF\_4)
\AB{reveals} a clear improvement over the baseline Mean SW except for the mIoU of the 
\AB{\texttt{rain} domain.} 
\AB{Using more} than two weights 
\AB{slightly improves predictive performance and calibration  almost consistently.}
The comparison with 
\AB{DE} shows a significantly better KL divergence and a partially better ECE. Only when comparing the mIoU of 
\AB{DE variant} with 4 members (DE\_4), 
\AB{WF\_2 to WF\_4 fare worse}.
\AB{We emphasize that the runtime of DE} 
increases linearly with the number of members. Thus, 
\AB{DE\_4 is 4$\times$ slower at inference than SWA and WF.} 


All in all, weight fusion has advantages over its main competitor SWA in terms of performance and calibration. Especially when it comes to generalization capability, WF excels. The comparison to 
\AB{DE} shows that weight fusion leads to a better calibration and can keep up with the performance of the 
an ensemble 
\AB{of three networks}, 
\AB{while taking} only 1/3 of the runtime in this case. Additional visual results and per class IoU analysis for the BDD test data are included in Appendix~\ref{more_visual_results}.

\begin{table}[t]
\caption{Results of weight fusion and comparable methods on BDD100K (in-distribution) and ACDC (out-of-distribution). Please refer to \S\ref{sec:results_method_decription} for a full explanation of the methods.}
\label{tab:overall_results}
\scriptsize
 \centering
\resizebox{\textwidth}{!}{%
\begin{tabular}{c|ccc|ccc|ccc|ccc|ccc}
\toprule
                                   & \multicolumn{3}{c|}{BDD100K}                                        & \multicolumn{12}{c}{ACDC}               \\ \hline
\toprule
 & \multicolumn{3}{c|}{test data} & \multicolumn{3}{|c|}{fog} & \multicolumn{3}{|c|}{night} & \multicolumn{3}{|c|}{rain} & \multicolumn{3}{|c|}{snow} \\ \hline

\multicolumn{1}{r|}{Methods} & mIoU (\%) ($\uparrow$) & ECE ($\downarrow$) & \multicolumn{1}{c|}{KL ($\uparrow$)} & mIoU (\%) ($\uparrow$) & ECE ($\downarrow$) & KL ($\uparrow$)& mIoU (\%) & ECE ($\downarrow$) & KL ($\uparrow$)& mIoU (\%) ($\uparrow$) & ECE ($\downarrow$) & KL ($\uparrow$)& mIoU (\%) & ECE ($\downarrow$) & KL ($\uparrow$)\\ \hline
\multicolumn{1}{r|}{Single weight (SW)} & 60.81 & 0.246 & 0.4222 & 44.34 & 0.191 & 0.694 & 22.08 & 0.214 & 0.869 & 42.08 & 0.189 & 0.792 & 37.28 & 0.206 & 0.622       \\
\multicolumn{1}{r|}{SWA} &  61.67 \color{ForestGreen}(+0.86) & 0.111 \color{ForestGreen}(-0.135) & 0.944 \color{ForestGreen}(+0.522) & 44.57 \color{ForestGreen}(+0.23) & 0.139 \color{ForestGreen}(-0.052) & 0.823 \color{ForestGreen}(+0.129) & 22.80 \color{ForestGreen}(+0.72) & 0.169 \color{ForestGreen}(-0.045)& 0.830 \color{red}(-0.039) & 42.25 \color{ForestGreen}(+0.17) & 0.138 \color{ForestGreen}(-0.051) & 0.864 \color{ForestGreen}(+0.072) & 36.94 \color{red}(-0.34) & 0.17 \color{ForestGreen}(-0.036) & 0.888 \color{ForestGreen}(+0.266)  \\
\multicolumn{1}{r|}{WF (ours)} & 62.05 \color{ForestGreen}(+1.24) & 0.080 \color{ForestGreen}(-0.166) & 1.103 \color{ForestGreen}(+0.681) & 47.38 \color{ForestGreen}(+3.04) & 0.149 \color{ForestGreen}(-0.042) & 1.135 \color{ForestGreen}(+0.441) & 24.95 \color{ForestGreen}(+2.87) & 0.159 \color{ForestGreen}(-0.055) & 0.863 \color{red}(-0.006) & 44.60 \color{ForestGreen}(+2.52) & 0.141 \color{ForestGreen}(-0.048) & 1.051 \color{ForestGreen}(+0.259) & 41.56 \color{ForestGreen}(+4.28) & 0.162 \color{ForestGreen}(-0.044) & 1.018 \color{ForestGreen}(+0.396) \\ \hline \hline
\multicolumn{1}{r|}{Mean SW} & 61.14 & 0.194 & 0.720 & 45.48 & 0.198 & 0.729 & 23.32 & 0.231 & 0.646 & 41.99 & 0.198 & 0.837 & 38.45 & 0.214 & 0.656 \\
\multicolumn{1}{r|}{DE\_2} & 62.04 \color{ForestGreen}(+0.90) & 0.149 \color{ForestGreen}(-0.045) & 0.890 \color{ForestGreen}(+0.17) & 42.29 \color{red}(-3.19) & 0.239 \color{red}(+0.041) & 0.688 \color{red}(-0.041) & 21.66 \color{red}(-1.66) & 0.269 \color{red}(+0.038) & 0.507 \color{red}(-0.139) & 38.80 \color{red}(-3.19) & 0.232 \color{red}(+0.034) & 0.868 \color{ForestGreen}(+0.031) & 35.53 \color{red}(-2.92) & 0.249 \color{red}(+0.035) & 0.566 \color{red}(-0.09)   \\
\multicolumn{1}{r|}{DE\_3} & 62.39 \color{ForestGreen}(+1.25) & 0.144 \color{ForestGreen}(-0.05) & 0.912 \color{ForestGreen}(+0.192) & 46.16 \color{ForestGreen}(+0.68) & 0.178 \color{ForestGreen}(-0.02) & 0.689 \color{red}(-0.04) & 24.12 \color{ForestGreen}(+0.8) & 0.210 \color{ForestGreen}(-0.021) & 0.550 \color{red}(-0.139) & 42.68 \color{ForestGreen}(+0.69) & 0.183 \color{ForestGreen}(-0.015) & 0.798 \color{red}(-0.039) & 38.51 \color{ForestGreen}(+0.06) & 0.202 \color{ForestGreen}(-0.012) & 0.546 \color{red}(-0.11)\\
\multicolumn{1}{r|}{DE\_4} & 62.88 \color{ForestGreen}(+1.74) & 0.159 \color{ForestGreen}(-0.035) & 0.883 \color{ForestGreen}(+0.163) & 47.35 \color{ForestGreen}(+1.87) & 0.170 \color{ForestGreen}(-0.028) & 0.807 \color{ForestGreen}(+0.078) & 24.78 \color{ForestGreen}(+1.46) & 0.192 \color{ForestGreen}(-0.039) & 0.470 \color{red}(-0.176) & 43.40 \color{ForestGreen}(+1.41) & 0.159 \color{ForestGreen}(-0.039) & 0.736 \color{red}(-0.101) & 39.61 \color{ForestGreen}(+1.16) & 0.187 \color{ForestGreen}(-0.027) & 0.568 \color{red}(-0.088)\\
\multicolumn{1}{r|}{WF\_2 (ours)} & 62.17 \color{ForestGreen}(+1.03) & 0.130 \color{ForestGreen}(-0.064) & 1.106 \color{ForestGreen}(+0.386) & 46.58 \color{ForestGreen}(+1.10) & 0.185 \color{ForestGreen}(-0.013) & 1.461 \color{ForestGreen}(+0.732) & 23.78 \color{ForestGreen}(+0.46) & 0.189 \color{ForestGreen}(-0.042) & 1.534 \color{ForestGreen}(+0.888) & 41.63 \color{red}(-0.36) & 0.189 \color{ForestGreen}(-0.009)& 1.310 \color{ForestGreen}(+0.473) & 38.84 \color{ForestGreen}(+0.39) & 0.207 \color{ForestGreen}(-0.007) & 1.299 \color{ForestGreen}(+0.643) \\ 
\multicolumn{1}{r|}{WF\_3 (ours)} & 62.27 \color{ForestGreen}(+1.13) & 0.131 \color{ForestGreen}(-0.063) & 1.113 \color{ForestGreen}(+0.393) & 46.73 \color{ForestGreen}(+1.25) & 0.193 \color{ForestGreen}(-0.005) & 1.647 \color{ForestGreen}(+0.918) & 23.91 \color{ForestGreen}(+0.59) & 0.192 \color{ForestGreen}(-0.022) & 1.610 \color{ForestGreen}(+0.964) & 41.47 \color{red}(-0.52) & 0.189 \color{ForestGreen}(-0.009) & 1.340 \color{ForestGreen}(+0.503) & 38.81 \color{ForestGreen}(+0.36) & 0.210 \color{red}(+0.004) & 1.290 \color{ForestGreen}(+0.668)\\
\multicolumn{1}{r|}{WF\_4 (ours)} & 62.32 \color{ForestGreen}(+1.18) & 0.114 \color{ForestGreen}(-0.08) & 1.130 \color{ForestGreen}(+0.41) & 46.80 \color{ForestGreen}(+1.32) & 0.164 \color{ForestGreen}(-0.027) & 1.578 \color{ForestGreen}(+0.849) & 24.14 \color{ForestGreen}(+0.82) & 0.158 \color{ForestGreen}(-0.073) & 1.514 \color{ForestGreen}(+0.868) & 41.86 \color{red}(-0.13) & 0.177\color{ForestGreen}(-0.021) & 1.378 \color{ForestGreen}(+0.541) & 38.76 \color{ForestGreen}(+0.31) & 0.182 \color{ForestGreen}(-0.032) & 1.292 \color{ForestGreen}(+0.67)  

\end{tabular}}
\end{table}

\section{Conclusion}
We have presented a strategy to fuse two or more 
\AB{sets of weights into a single one.}
\AB{The weights can be computed by training from scratch or by finetuning.} We have derived from extensive experiments that the similarity in weight space between the weights must be as large as possible and at the same time the similarity in functional space must be as small as possible. To measure the functional space, we introduced a new testing method called oralce testing. 
\AB{When computing the weights by finetuning, }the validation loss plays a decisive role, which should be as close as possible to the local minimum. Based on 
\AB{extensive} studies with SoTA architectures (CNNs and Transformers) we presented the improvements in predictive performance and calibration. 
We have shown that equal-weighted fusion as performed in SWA does not usually lead to the best results and can even be detrimental if the cosine similarity is too low. 
We outperformed stochastic weight averaging (SWA) on the BDD100K test data as well as under presence of a distribution shift on the ACDC data in performance and calibration. Furthermore, comparison with deep ensembles shows that our weight fusion has slightly better calibration and comparable performance to the deep ensemble with 3 members, albeit the latter requires 
\AB{3$\times$} the runtime.

\section{Outlook}
(i) Considering the beneficial effect of our weight fusion, the question arises about the potential of more complex variants, such as layer-level fusion or nonlinear weight fusion. 
(ii) The question arises whether with a new loss function in training the weights can be driven further apart in their functional space while keeping the distance in weight space sufficiently small, which is expected to result in further improvement in performance and calibration. 
(iii) Based on the results in Fig.~\ref{fig:lr_rate_schedule} subplot 1 where the fused mIoU did not occur at the end of the cycle but during the cycle, the question arises whether SWA and its variants should better average the weights in the middle of the cycle than at the end.
(iv) Furthermore, we leave the applicability of weight fusion to other autonomous driving tasks like object detection or point cloud segmentation to future work.

\begin{ack}
The research leading to these results is funded by the Federal Ministry for Economic Affairs and Energy within the project "KI Absicherung - Safe AI for automated driving". The authors would like to thank the consortium for the successful cooperation.
\end{ack}

{
\small

\bibliographystyle{splncs04}
\bibliography{egbib}

\begin{thebibliography}{10}
\providecommand{\url}[1]{\texttt{#1}}
\providecommand{\urlprefix}{URL }
\providecommand{\doi}[1]{https://doi.org/#1}

\bibitem{aksela2003comparison}
Aksela, M.: Comparison of classifier selection methods for improving committee
  performance. In: International Workshop on Multiple Classifier Systems (2003)

\bibitem{ashukha2020pitfalls}
Ashukha, A., Lyzhov, A., Molchanov, D., Vetrov, D.: Pitfalls of in-domain
  uncertainty estimation and ensembling in deep learning. In: ICLR (2020)

\bibitem{cha2021swad}
Cha, J., Chun, S., Lee, K., Cho, H.C., Park, S., Lee, Y., Park, S.: Swad:
  Domain generalization by seeking flat minima. Advances in Neural Information
  Processing Systems  \textbf{34} (2021)

\bibitem{chen2018encoder}
Chen, L.C., Zhu, Y., Papandreou, G., Schroff, F., Adam, H.: Encoder-decoder
  with atrous separable convolution for semantic image segmentation. In:
  Proceedings of the European conference on computer vision (ECCV). pp.
  801--818 (2018)

\bibitem{Cheng_2020_CVPR}
Cheng, B., Collins, M.D., Zhu, Y., Liu, T., Huang, T.S., Adam, H., Chen, L.C.:
  Panoptic-deeplab: A simple, strong, and fast baseline for bottom-up panoptic
  segmentation. In: CVPR (2020)

\bibitem{cordts2016cityscapes}
Cordts, M., Omran, M., Ramos, S., Rehfeld, T., Enzweiler, M., Benenson, R.,
  Franke, U., Roth, S., Schiele, B.: The cityscapes dataset for semantic urban
  scene understanding. In: Proceedings of the IEEE conference on computer
  vision and pattern recognition. pp. 3213--3223 (2016)

\bibitem{dehghani2021benchmark}
Dehghani, M., Tay, Y., Gritsenko, A.A., Zhao, Z., Houlsby, N., Diaz, F.,
  Metzler, D., Vinyals, O.: The benchmark lottery. arXiv preprint
  arXiv:2107.07002  (2021)

\bibitem{dinh2017sharp}
Dinh, L., Pascanu, R., Bengio, S., Bengio, Y.: Sharp minima can generalize for
  deep nets. In: International Conference on Machine Learning. pp. 1019--1028.
  PMLR (2017)

\bibitem{dosovitskiy2021image}
Dosovitskiy, A., Beyer, L., Kolesnikov, A., Weissenborn, D., Zhai, X.,
  Unterthiner, T., Dehghani, M., Minderer, M., Heigold, G., Gelly, S.,
  Uszkoreit, J., Houlsby, N.: An image is worth 16x16 words: Transformers for
  image recognition at scale. In: ICLR (2021)

\bibitem{fort2019deep}
Fort, S., Hu, H., Lakshminarayanan, B.: Deep ensembles: A loss landscape
  perspective. arXiv preprint arXiv:1912.02757  (2019)

\bibitem{franchi2021robust}
Franchi, G., Belkhir, N., Ha, M.L., Hu, Y., Bursuc, A., Blanz, V., Yao, A.:
  Robust semantic segmentation with superpixel-mix. arXiv preprint
  arXiv:2108.00968  (2021)

\bibitem{gal2016dropout}
Gal, Y., Ghahramani, Z.: Dropout as a bayesian approximation: Representing
  model uncertainty in deep learning. In: ICML (2016)

\bibitem{garipov2018loss}
Garipov, T., Izmailov, P., Podoprikhin, D., Vetrov, D.P., Wilson, A.G.: Loss
  surfaces, mode connectivity, and fast ensembling of dnns. Advances in neural
  information processing systems  \textbf{31} (2018)

\bibitem{guo2017calibration}
Guo, C., Pleiss, G., Sun, Y., Weinberger, K.Q.: On calibration of modern neural
  networks. In: International Conference on Machine Learning. pp. 1321--1330.
  PMLR (2017)

\bibitem{guo2022stochastic}
Guo, H., Jin, J., Liu, B.: Stochastic weight averaging revisited. arXiv
  preprint arXiv:2201.00519  (2022)

\bibitem{gustafsson2020evaluating}
Gustafsson, F.K., Danelljan, M., Sch{\"o}n, T.B.: Evaluating scalable bayesian
  deep learning methods for robust computer vision. In: CVPRW (2020)

\bibitem{huang2017snapshot}
Huang, G., Li, Y., Pleiss, G., Liu, Z., Hopcroft, J.E., Weinberger, K.Q.:
  Snapshot ensembles: Train 1, get m for free (2017)

\bibitem{izmailov2018averaging}
Izmailov, P., Podoprikhin, D., Garipov, T., Vetrov, D., Wilson, A.G.: Averaging
  weights leads to wider optima and better generalization. arXiv preprint
  arXiv:1803.05407  (2018)

\bibitem{kullback1951information}
Kullback, S., Leibler, R.A.: On information and sufficiency. The annals of
  mathematical statistics  \textbf{22}(1),  79--86 (1951)

\bibitem{lakshminarayanan2017simple}
Lakshminarayanan, B., Pritzel, A., Blundell, C.: Simple and scalable predictive
  uncertainty estimation using deep ensembles. In: NeurIPS (2017)

\bibitem{lee2015m}
Lee, S., Purushwalkam, S., Cogswell, M., Crandall, D., Batra, D.: Why m heads
  are better than one: Training a diverse ensemble of deep networks. arXiv
  preprint arXiv:1511.06314  (2015)

\bibitem{lin2017feature}
Lin, T.Y., Doll{\'a}r, P., Girshick, R., He, K., Hariharan, B., Belongie, S.:
  Feature pyramid networks for object detection. In: CVPR (2017)

\bibitem{long2015fully}
Long, J., Shelhamer, E., Darrell, T.: Fully convolutional networks for semantic
  segmentation. In: CVPR (2015)

\bibitem{mukhoti2020calibrating}
Mukhoti, J., Kulharia, V., Sanyal, A., Golodetz, S., Torr, P., Dokania, P.:
  Calibrating deep neural networks using focal loss. Advances in Neural
  Information Processing Systems  \textbf{33},  15288--15299 (2020)

\bibitem{neuhold2017mapillary}
Neuhold, G., Ollmann, T., Rota~Bulo, S., Kontschieder, P.: The mapillary vistas
  dataset for semantic understanding of street scenes. In: ICCV (2017)

\bibitem{neumann2018relaxed}
Neumann, L., Zisserman, A., Vedaldi, A.: Relaxed softmax: Efficient confidence
  auto-calibration for safe pedestrian detection  (2018)

\bibitem{ovadia2019can}
Ovadia, Y., Fertig, E., Ren, J., Nado, Z., Sculley, D., Nowozin, S., Dillon,
  J., Lakshminarayanan, B., Snoek, J.: Can you trust your model's uncertainty?
  evaluating predictive uncertainty under dataset shift. In: NeurIPS (2019)

\bibitem{paszke2016enet}
Paszke, A., Chaurasia, A., Kim, S., Culurciello, E.: Enet: A deep neural
  network architecture for real-time semantic segmentation. arXiv preprint
  arXiv:1606.02147  (2016)

\bibitem{platt1999probabilistic}
Platt, J., et~al.: Probabilistic outputs for support vector machines and
  comparisons to regularized likelihood methods. Advances in large margin
  classifiers  \textbf{10}(3),  61--74 (1999)

\bibitem{rame2021dice}
Rame, A., Cord, M.: Dice: Diversity in deep ensembles via conditional
  redundancy adversarial estimation. In: ICLR (2021)

\bibitem{Ronneberger2015UNet}
Ronneberger, O., Fischer, P., Brox, T.: U-net: Convolutional networks for
  biomedical image segmentation. In: MICCAI (2015)

\bibitem{sakaridis2021acdc}
Sakaridis, C., Dai, D., Van~Gool, L.: Acdc: The adverse conditions dataset with
  correspondences for semantic driving scene understanding. In: Proceedings of
  the IEEE/CVF International Conference on Computer Vision. pp. 10765--10775
  (2021)

\bibitem{strudel2021segmenter}
Strudel, R., Garcia, R., Laptev, I., Schmid, C.: Segmenter: Transformer for
  semantic segmentation. In: Proceedings of the IEEE/CVF International
  Conference on Computer Vision. pp. 7262--7272 (2021)

\bibitem{tarvainen2017mean}
Tarvainen, A., Valpola, H.: Mean teachers are better role models:
  Weight-averaged consistency targets improve semi-supervised deep learning
  results. Advances in neural information processing systems  \textbf{30}
  (2017)

\bibitem{thulasidasan2019mixup}
Thulasidasan, S., Chennupati, G., Bilmes, J.A., Bhattacharya, T., Michalak, S.:
  On mixup training: Improved calibration and predictive uncertainty for deep
  neural networks. Advances in Neural Information Processing Systems
  \textbf{32} (2019)

\bibitem{wang2020deep}
Wang, J., Sun, K., Cheng, T., Jiang, B., Deng, C., Zhao, Y., Liu, D., Mu, Y.,
  Tan, M., Wang, X., et~al.: Deep high-resolution representation learning for
  visual recognition. IEEE transactions on pattern analysis and machine
  intelligence  \textbf{43}(10),  3349--3364 (2020)

\bibitem{wen2020batchensemble}
Wen, Y., Tran, D., Ba, J.: Batchensemble: an alternative approach to efficient
  ensemble and lifelong learning. In: iclr (2020)

\bibitem{xie2021segformer}
Xie, E., Wang, W., Yu, Z., Anandkumar, A., Alvarez, J.M., Luo, P.: Segformer:
  Simple and efficient design for semantic segmentation with transformers.
  arXiv preprint arXiv:2105.15203  (2021)

\bibitem{yu2017dilated}
Yu, F., Koltun, V., Funkhouser, T.: Dilated residual networks. In: CVPR (2017)

\bibitem{yu2018bdd100k}
Yu, F., Xian, W., Chen, Y., Liu, F., Liao, M., Madhavan, V., Darrell, T.:
  Bdd100k: A diverse driving video database with scalable annotation tooling.
  arXiv preprint arXiv:1805.04687  \textbf{2}(5), ~6 (2018)

\bibitem{yun2019cutmix}
Yun, S., Han, D., Oh, S.J., Chun, S., Choe, J., Yoo, Y.: Cutmix: Regularization
  strategy to train strong classifiers with localizable features. In:
  Proceedings of the IEEE/CVF international conference on computer vision. pp.
  6023--6032 (2019)

\bibitem{zhang2017mixup}
Zhang, H., Cisse, M., Dauphin, Y.N., Lopez-Paz, D.: mixup: Beyond empirical
  risk minimization. arXiv preprint arXiv:1710.09412  (2017)

\bibitem{zhao2018icnet}
Zhao, H., Qi, X., Shen, X., Shi, J., Jia, J.: Icnet for real-time semantic
  segmentation on high-resolution images. In: ECCV (2018)

\bibitem{zhao2017pyramid}
Zhao, H., Shi, J., Qi, X., Wang, X., Jia, J.: Pyramid scene parsing network.
  In: CVPR (2017)

\bibitem{zheng2021rethinking}
Zheng, S., Lu, J., Zhao, H., Zhu, X., Luo, Z., Wang, Y., Fu, Y., Feng, J.,
  Xiang, T., Torr, P.H., Zhang, L.: Rethinking semantic segmentation from a
  sequence-to-sequence perspective with transformers. In: CVPR (2021)

\bibitem{ade20k}
Zhou, B., Zhao, H., Puig, X., Fidler, S., Barriuso, A., Torralba, A.: Scene
  parsing through ade20k dataset. In: Proceedings of the IEEE Conference on
  Computer Vision and Pattern Recognition (2017)

\end{thebibliography}

}


\appendix

\setcounter{section}{0}
\renewcommand\thesection{\Alph{section}}
\renewcommand\thesubsection{\thesection.\Alph{subsection}}
\section{Related Work}

\parag{Ensemble strategies.} Ensembles are a simple approach for generating robust predictions by averaging decisions from multiple networks trained with different random initialization seeds. Deep Ensemble (DE)~\cite{lakshminarayanan2017simple} is a highly effective method for achieving both high accuracy and calibration scores~\cite{ovadia2019can,gustafsson2020evaluating,fort2019deep}. However, its computational costs for training and inferences make them prohibitive \AB{for real-time predictions}. These limitations have been addressed by several approaches that generate an ensemble from a single training run by collecting intermediate checkpoints~\cite{huang2017snapshot,garipov2018loss}, imbuing a network with multiple prediction heads~\cite{lee2015m} or multiple low-rank weights~\cite{wen2020batchensemble}, using multiple Dropout masks during inference~\cite{gal2016dropout}, multiple test-time augmentations of an input image~\cite{ashukha2020pitfalls}. These approaches, while faster to train, still have slow runtime due to the multiple forward passes that are performed. In practical settings, fast predictions are key and our approach does not bring any extra compute cost at test-time. 

\parag{Fusing strategies.} Deep Ensemble~\cite{lakshminarayanan2017simple} can naturally fuse predictions from the individual networks by averaging predictions. In the quest of mimicking the robustness of ensembles and of flat minima~\cite{dinh2017sharp} that generalize better, some approaches average network weights instead~\cite{tarvainen2017mean,izmailov2018averaging,cha2021swad,guo2022stochastic}. Most approaches propose simple averaging of the network weights. Here we show that simple averaging~\cite{izmailov2018averaging} is not an optimal strategy and that the way in which model weights are fused can greatly impact the final performance of a model.

\parag{Calibration.} Calibration is another key requirement for models deployed in real-world settings as it facilitates thresholding of confidence scores from network predictions. However, DNNs tend to be overconfident even when they are incorrect~\cite{guo2017calibration}. This limitation can be addressed by a post-training operation that learns to map confidence scores to probabilities, e.g., using Platt scaling~\cite{platt1999probabilistic}, temperature scaling~\cite{guo2017calibration}. However these strategies do not generalize well under distribution shift~\cite{ovadia2019can} as the scaling coefficient can overfit to the \AB{in-distribution} data. Alternative approaches propose strategies for improving the calibration of the DNNs during training. Data augmentation strategies based on mixing~\cite{zhang2017mixup,yun2019cutmix} have been shown to improve calibration~\cite{thulasidasan2019mixup,franchi2021robust} along with crafted loss functions~\cite{mukhoti2020calibrating}. WF also aims for improved calibration but does not require crafted loss functions or augmentation strategies and generalizes across network architectures.

\parag{Semantic segmentation.} Semantic segmentation is a key visual perception task, in particular for autonomous driving~\cite{cordts2016cityscapes, yu2018bdd100k, neuhold2017mapillary} where it enables an understanding of the environment surrounding the vehicle. The most effective semantic segmentation architectures are currently based on fully convolutional DNNs~\cite{long2015fully} with encoder-decoder structures. In the last few years tremendous amount of architectures for semantic segmentation have been proposed aiming for fine high-resolution predictions~\cite{Ronneberger2015UNet,  chen2018encoder,yu2017dilated,wang2020deep,Cheng_2020_CVPR}, multi-scale processing~\cite{zhao2017pyramid, lin2017feature} or computational efficiency~\cite{paszke2016enet,zhao2018icnet}. Furthermore, the recent progress in Vision Transformers~\cite{dosovitskiy2021image} has facilitated the rise of a new generation of transformer-based decoders for semantic segmentation~\cite{strudel2021segmenter,xie2021segformer,zheng2021rethinking}. We consider here prominent convolutional and transformer architectures (DeepLabV3+\cite{chen2018encoder}, HRNet\cite{wang2020deep}, Segmenter\cite{strudel2021segmenter}) to illustrate the generalization of our WF strategy.

\section{Two Further Findings}
\label{further_finding}
\subsection*{Finding 4: If the weights to be fused have a very high similarity in weight space, the weighting parameter (alpha) plays a minor role in the fusion.}
$\bar{\theta} =  \alpha \theta^{(1)} + (1-\alpha) \theta^{(2)}$, so if $\theta^{(1)} \sim \theta^{(2)}$, then $\bar{\theta} \sim \theta^{(1)} \sim \theta^{(2)}$, thus 
\AB{$\alpha$} plays a minor role. 
A high cosine similarity leads to a flat course of the mIoU along the 
\AB{$\alpha$} parameters (see red and orange curve in Fig.~\ref{fig:finetuning_experiments}(c)). Equally weighted fusion is only beneficial when cosine similarity is very high. 

\begin{figure}[t]
    \includegraphics[width=1.0\textwidth]{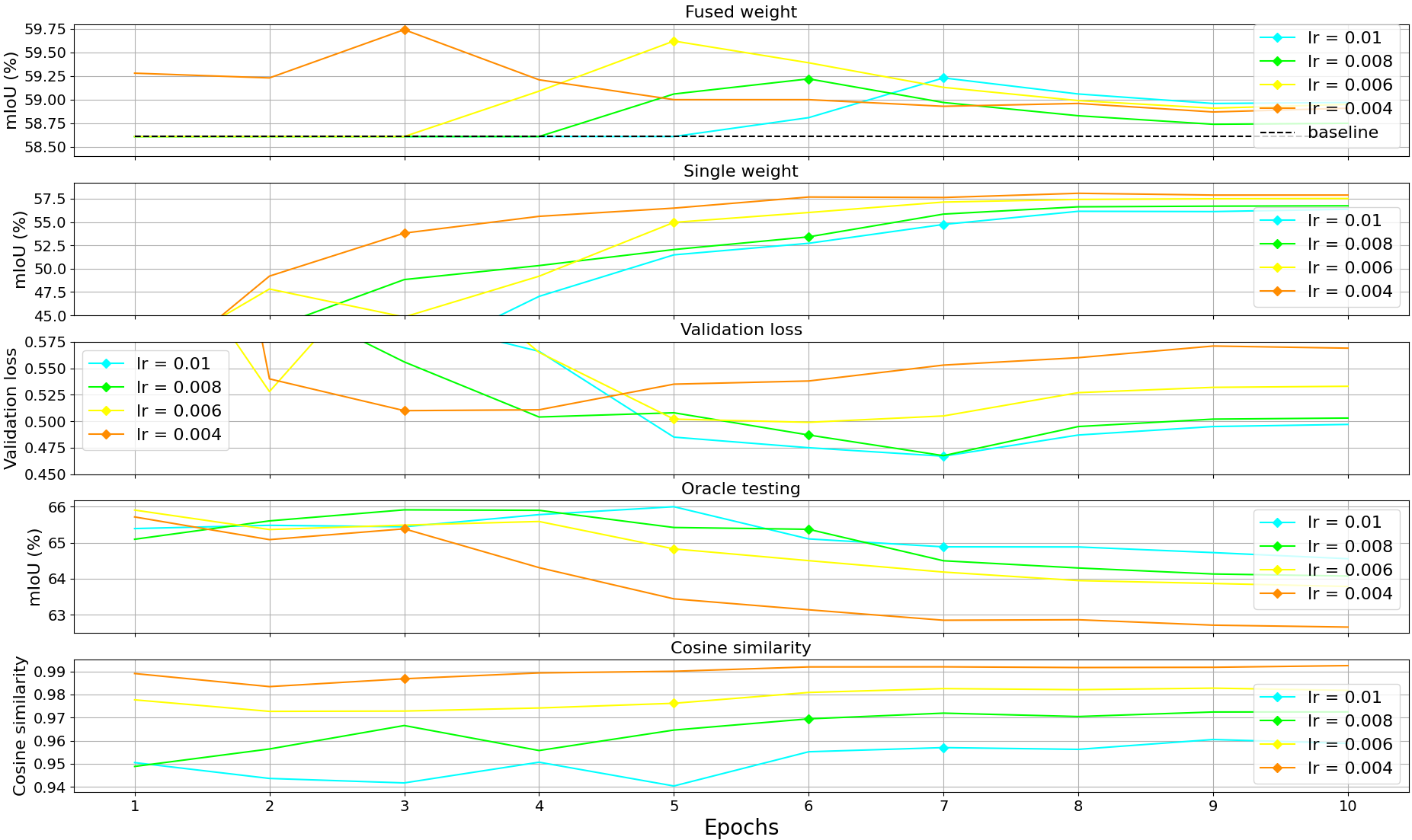}
	\centering
	\caption{
	Results for the 10 epochs finetuning with the cyclic learning rate schedule. The first and second subplots show the mIoU for the fused and finetuned weights, respectively. The third subplot shows the validation loss for the finetuned weights. The fourth and fifth subplots show the oracle testing and cosine similarity results, respectively, using the starting weight and the finetuned weights.
	}
	\label{fig:lr_rate_schedule}
\end{figure}

The next finding refers to Fig.~\ref{fig:lr_rate_schedule}, which is used to analyze the requirements for the weights along the 10-epoch finetuning. For the sake of clarity, we restrict ourselves to learning rates 0.004 to 0.01. The evaluations include the mIoU of the fused weight, the mIoU of the single weight of each epoch, the validation loss, the oracle testing, and the cosine similarity. Please note that for the three subplots fused weight, oracle testing and cosine similarity, the starting weight and the weight of the respective epoch was used. The parameter 
\AB{$\alpha$} used for the weight fusion was determined for each epoch using grid search in 0.05 steps. 
Based on the first subplot in Fig.~\ref{fig:lr_rate_schedule} it can be seen that the fusion of the starting weight with the weight after epoch 3, 5, 6 and 7 respectively reaches the highest mIoU value. 
For easier comparison of the subplots for these epochs, they are marked with a diamond marker. 
From Fig.~\ref{fig:lr_rate_schedule} we derive the following finding:
\subsection*{Finding 5: A successful weight fusion is strongly correlated with the validation loss of the second weight in this finetuning setup.}
\AB{In the subplot of the validation loss, we observe that the diamond markers are mostly in the minimum.}
This is not completely true for the learning rate 0.008, because here the minimum is one epoch further at epoch 7. However, 
\AB{when checking} the oracle test, there is a reduction of the mIoU from epoch 6 to 7. We see this as confirmation of another claim that a high mIoU in oracle testing is important as well.


\section{WF and Cosine Similarity Pytorch Code Snippet}\label{pytorch_code_snippet}
The code for WF for two weights is listetd in~\ref{lst:weightfusion} and the code for calculating cosine similarity for two weights is listed in~\ref{lst:cosine}. For more details, please refer to the captions and comments of the code snippets.

\begin{lstlisting}[basicstyle=\small, language=Python, label={lst:weightfusion}, caption={We iterate over the layers and calculate the fused weight (checkpoint\_fused).}, captionpos=b]
for k in checkpoint[0]['state_dict'].keys():
    checkpoint_fused['state_dict'][k] = \
        alpha * checkpoint[0]['state_dict'][k] + \ 
        beta * checkpoint[1]['state_dict'][k]
\end{lstlisting}

\begin{minipage}{\linewidth}
\begin{lstlisting}[basicstyle=\small, language=Python, label={lst:cosine}, caption={We iterate over the layers and flatten the parameters of each layer. We then concatenate the parameters before calculating the cosine similarity.},captionpos=b]
 cos = nn.CosineSimilarity(dim=1, eps=1e-6)
 # iterate over the layers and concatenate the parameters
 for k in checkpoint[0]['state_dict'].keys():
     # remove num_batches_tracked in bn layers
     if len(list(checkpoint[0]['state_dict'][k].shape)) == 0:
         continue
     # flatten the parameters before concatenation
     vector_key1 = checkpoint[0]['state_dict'][k].view(1, -1)
     vector_key2 = checkpoint[1]['state_dict'][k].view(1, -1)
     vector1 = torch.cat((vector_key1, vector1), 1)
     vector2 = torch.cat((vector_key2, vector2), 1)
 cos_sim = cos(vector1, vector2)
\end{lstlisting}
\end{minipage}

\section{Datasets}\label{adapt_bdd}
In the following, we describe the datasets used and specifically address the adaptations made on the BDD100K dataset. All three datasets contain the same 19 classes and follow the Cityscapes label protocol, allowing for easy transferability of models. 

\textbf{BDD100K.}  
BDD100K~\cite{yu2018bdd100k} is a large-scale and diverse dataset for autonomous driving. It contains diverse data related to weather conditions, environments and geographic. 
The size of the semantic segmentation part is 10k images, which is divided into 6k training, 1k validation and 3k test data with an resolution of 720x1280~pixels. Since the availability of the test dataset was not given at the time of the conducted experiments, we limited ourselves to the remaining 7k images. We split the 7k images again into 5k training, 1k validation and 1k test data. The 1k test data set corresponds to the originally designed 1k validation data set. The splitting into the 1k validation set was done randomly.
Due to numerous errors in the ground truth annotations and the resulting negative impact on result interpretations, we adapted the ground truth data. Obviously there was a systematic error in the BDD100K labeling tool, so that correctly segmented objects were assigned the wrong label. In these cases we changed the objects to void, which has the effect that these objects are ignored during training and evaluation. We made such a change in about every 5th image.

\textbf{Cityscapes.}  
As BDD100K, the Cityscapes dataset~\cite{cordts2016cityscapes} contains scenes from street traffic. The images were recorded in 50 cities of the countries Germany, France, Austria and Switzerland and have a resolution of 1024x2048~pixels. The 5000 image dataset is divided into 2975 training, 500 validation and 1525 test images. The label accuracy is very precise, which has made the dataset one of the most popular datasets for semantic segmentation, among others. Due to the limited accessibility of the test data set, we focus our experiments and evaluations on the validation data. 

\textbf{ACDC.}  
ACDC~\cite{sakaridis2021acdc} is a dataset for training and testing semantic segmentation models on road scenes with adverse visual conditions. The images were divided into the domains fog, night, rain, and snow and have a resolution of 1920x1080~pixe;s. Due to the easier accessibility, we use only the training dataset of these domains, each containing 400 images, to test the generalization ability of WF.

\setcounter{section}{2}


\section{Implementation details}\label{Used_hyperparameters}

In this section we share information about the details on the used DNN architectures and the used hyperparameters for the trainings. 

For DeepLabV3+~\cite{chen2018encoder}, we used the Resnet101 backbone pretrained on ImageNet and an out-stride of 16. 
 As for the Segmenter~\cite{strudel2021segmenter}, we used a Vision Transformer (ViT)~\cite{dosovitskiy2021image} backbone pretrained on ADE20K~\cite{ade20k} with patch size of 16 and token size of 768. 
For HRNet~\cite{wang2020deep}, we did not use the OCR module to upscale HRNet predictions for the sake of simplicity. Instead, we used simple bilinear interpolation. This is the reason why HRNet performs slightly worse than DeepLabV3+, even though HRNet usually performs better than DeepLabV3+, cf. Cityscapes benchmark. 

We provide the common hyper-parameters used for all three DNN architectures in Table~\ref{table:hyperparams}.

\begin{table}[t]
\renewcommand{\figurename}{Table}
\begin{center}
\scalebox{0.85}
{
\begin{tabular}{l|c}
\toprule
 {\textbf{Hyper-parameter}} &   \textbf{Value}   \\ 
\midrule
learning rate         & $0.01$   \\ 
\midrule
 learning rate schedule         & polynomial \\ 
 \midrule
loss        & cross-entropy  \\ 
 \midrule
optimizer     & SGD\\ 
 \midrule
 weight decay       & $0.0005$ \\ 
 \midrule
momentum  & $0.9$\\ 
\bottomrule
\end{tabular}
}
\end{center}
\caption{
{\textbf{Common hyper-parameter configuration used across experiments.}}
} 
\label{table:hyperparams}
\end{table}

Regarding the batch-size, this was chosen so that it could be done with one Quadro RTX 8000 that has 48 GB of GPU memory. Furthermore, we performed class weighting for cross-entropy to assign a larger weight to classes with a low pixel count.
To reduce the complexity of our experiments, we did not use data augmentation in the training.

The following mean and standard deviation were used for BDD100K and Cityscapes:

Cityscapes:
mean=(0.28689554, 0.32513303, 0.28389177), std=(0.18696375, 0.19017339, 0.18720214)

BDD:
mean=(0.3702, 0.4145, 0.4245), std=(0.057, 0.0658, 0.0755)

For the sake of reproducablity we share the github repositories where we took the code from:
\begin{itemize}
\item DeepLabV3+:
https://github.com/jfzhang95/pytorch-deeplab-xception
\item HRNet:
https://github.com/HRNet/HRNet-Semantic-Segmentation
\item Segmenter:
https://github.com/rstrudel/segmenter
\end{itemize}

\section{Additional Qualitative Results for WF}\label{Further_results}
\begin{figure}[t]
    \includegraphics[width=1\textwidth]{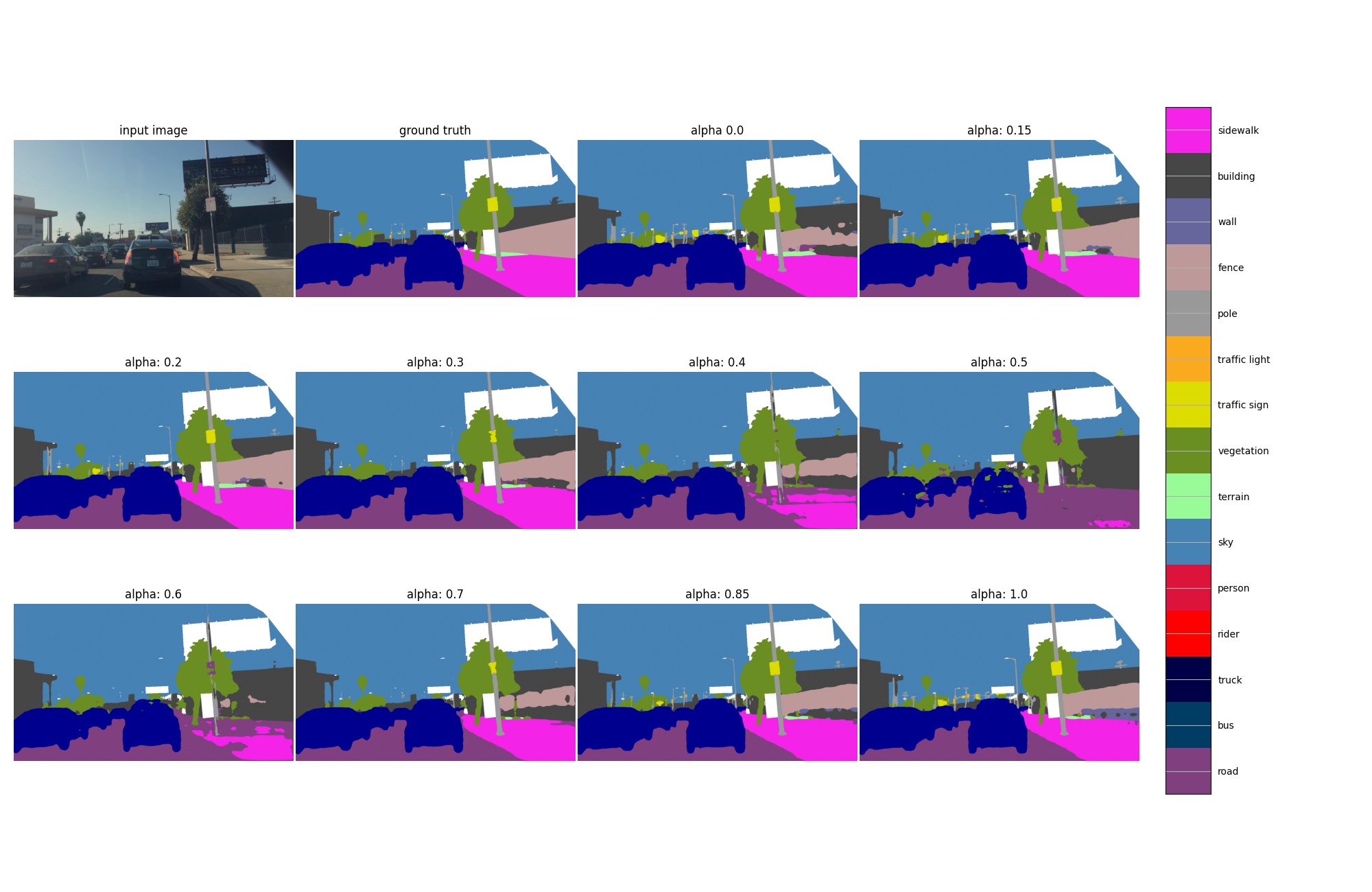}
	\centering
	\caption{Visual Semantic Segmentation results for DeepLabV3+ on BDD100K data. Please note that the color white refers to void.}
	\label{fig:f_r_2}
\end{figure}

\begin{figure}[h!]
    \includegraphics[width=1\textwidth]{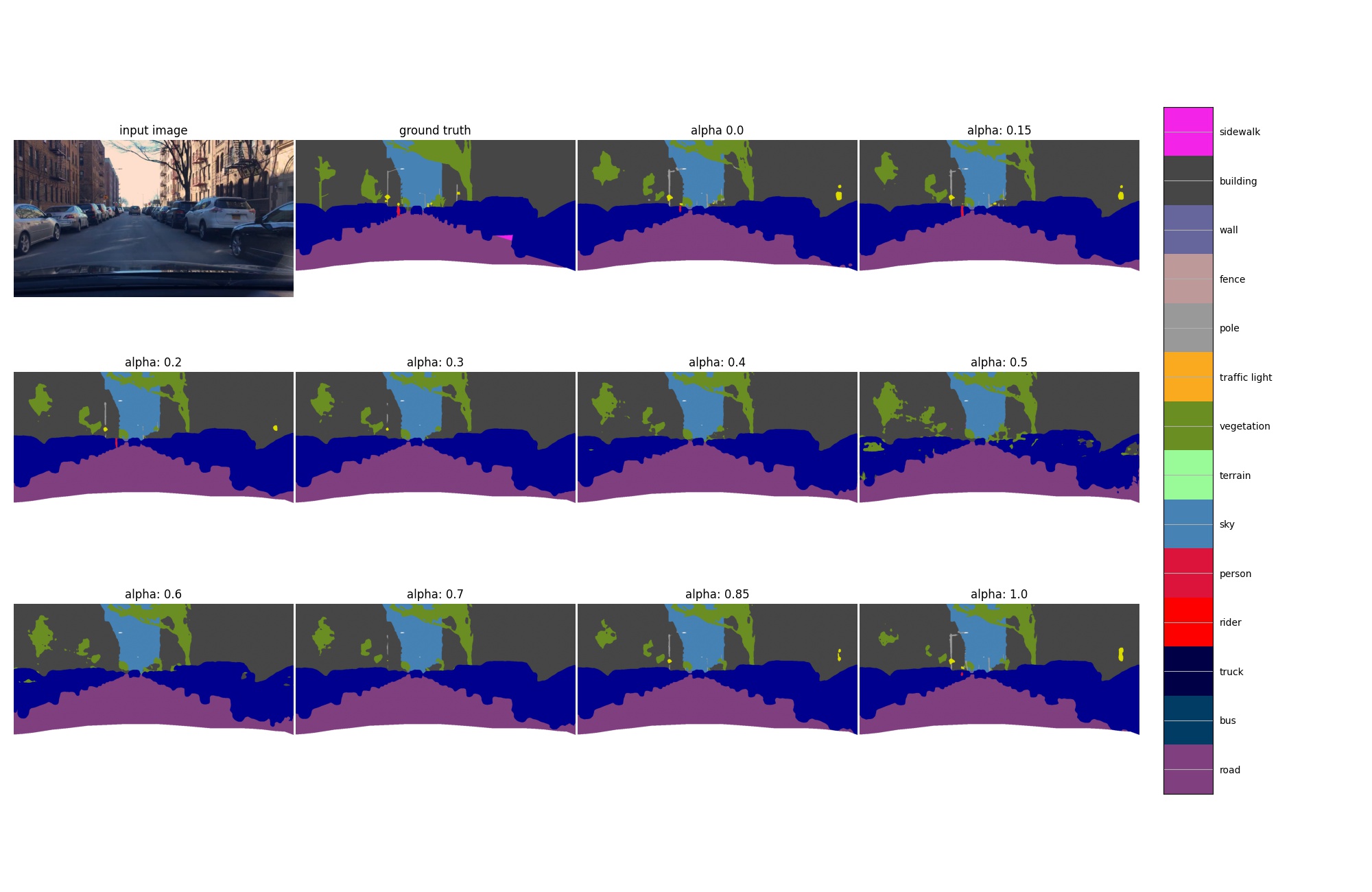}
	\centering
	\caption{Visual Semantic Segmentation results for DeepLabV3+ on BDD100K data. Please note that the color white refers to void.}
	\label{fig:f_r_3}
\end{figure}

In Figure~\ref{fig:f_r_2} and~\ref{fig:f_r_3} we show more visual results for WF with different fusion parameters alpha.
All images are from the BDD100K test dataset and the segmentation masks were generated using DeepLabV3+. 
Alpha=0.0 and alpha=1.0 refers to the single weights and serves as baselines to alpha=0.15, which according to 
Figure~1 in the main paper led to the best results.
We want to highlight in Figure~\ref{fig:f_r_2} the class \glqq fence\grqq{} on the right side of the image. The segmentation of \glqq fence\grqq{} is different from both baselines, making it clear that the fused weight is a new function and not just a visual combination of the segmentation masks of the baselines.
In Figure~\ref{fig:f_r_3}, we want to focus on the class \glqq person\grqq{} centered in the image. Only the fused weight with alpha=0.15 is able to segment \glqq person\grqq{} completely.

\section{Practical Advice when Applying WF to new Problems.}\label{Practical_Advice}

The goal is to find a good compromise between cosine similarity and oracle testing. This trade-off is to achieve the highest possible value for the oracle tests while keeping the cosine similarity not too low (in most of our experiments, a value higher than ~0.93 has proven to be effective).
How can we exploit this knowledge to generate appropriate weights for fusion? 

Broadly speaking: The higher the variance in training/finetuning the higher the value in oracle testing. At the same time, this reduces the cosine similarity. From this we conclude that the variance is the parameter for the right compromise. There are many ways to vary the variance in training/finetuning, e.g. by using a different learning rate, optimizer or initialization seed. By using two different learning rates for training from the scratch, the variance and thus the oracle testing can be increased. If one wants to reduce the variance, it is advisable to set all hyperparameters to the same value and only change the initialization seed. In our experiments it has been shown that the learning rate is a good parameter to adjust the variance during training and finetuning. 
Specifically, finetuning provides an easier and less computationally expensive way to generate the desired weights for fusion. Our results in terms of predictive performance and calibration show no disadvantages to training from the scratch.

How can grid search for alpha be made more computationally efficient?
After the appropriate weights have been generated, a grid search is applied to find the appropriate alpha parameter. The step size of 0.05 has proven to be useful, but this does not have to be run stupendously for all possibilities, since the changes of the mIoU are usually linear. I.e. if the alpha value of 0.05 as well as 0.1 provides an improvement, but 0.15 does not, the grid search can continue with alpha = 0.95. If no improvement can be measured at 0.95, the grid search can already be terminated. In this example, the best fused weight can be achieved with alpha 0.1.

\section{Analysis of WF on SWA-Setting}\label{further_analysis_wf}
\begin{figure}[h!]
    \includegraphics[width=1\textwidth]{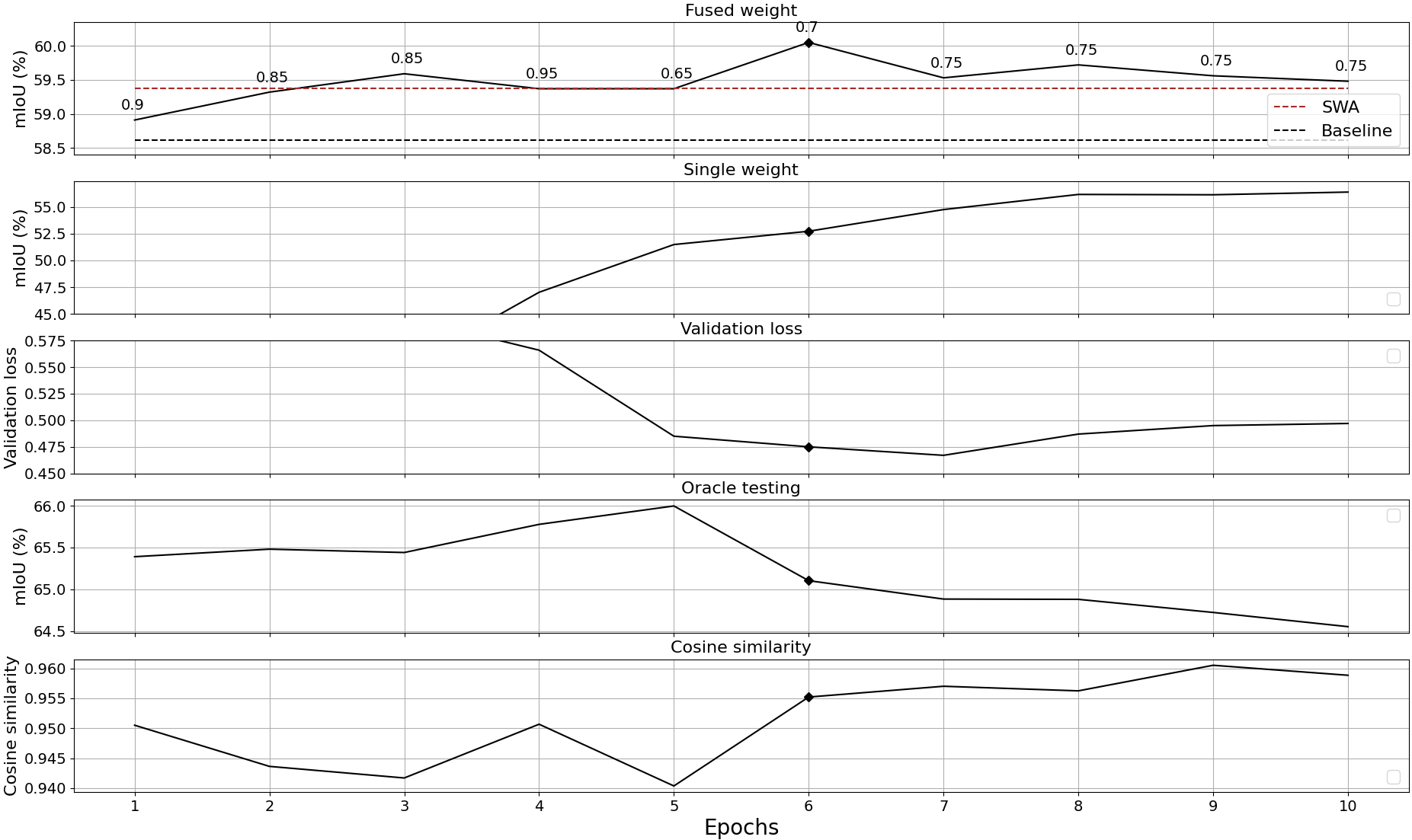}
	\centering
	\caption{Analysis of the WF on the BDD100K validation data. The first and second subplots show the mIoU for the fused and
finetuned weights, respectively. The third subplot shows the validation loss for
the finetuned weights. The fourth and fifth subplots show the oracle testing and
cosine similarity results, respectively, using the starting point and the weight of the corresponding epoch.}
	\label{fig:swa_comparison_with_baseline_supp}
\end{figure}

Figure~\ref{fig:swa_comparison_with_baseline_supp} shows the analysis of the WF on the BDD100K validation data which we use as a comparison to SWA, cf.~\ref{sec:results_method_decription}. 

The structure of the plot is the same as in 
Figure~5 from the main paper. The first subplot shows the mIoU after weight fusion of the starting weight with the weight of the respective epoch. The mIoU of the starting weight is called the baseline and the mIoU of SWA is also plotted for reference. The second and third subplots show the mIoU and validation loss of the weight of the corresponding epoch, respectively. Subplot 4 and 5 show the result of oracle testing and cosine similarity between the start weight and the weights of the corresponding epoch. The diamond marker indicates the highest mIoU of the fused weight, which is at epoch 6. At this point, the optimal compromise between low validation loss, high oracle testing and high cosine similarity is reached.

\section{Selection of the Weights for DE\_2, DE\_3, DE\_4, and WF\_2, WF\_3, WF\_4}\label{comparison_to_de}

The four weight files used in DE\_4 and WF\_4 were taken from subsection 4.1, in which DeepLabV3+ was trained four times from the scratch on BDD100K. 
The selection of weights for DE\_2, DE\_3 and WF\_2, WF\_3 was determined by evaluating the mIoU on the training data. The combination of weights (checkpoint IDs 1 to 4) that achieved the highest mIoU on the training data was selected for evaluation on the test data. 
The bold values in tables~\ref{tab:DE_Member_Selection2} and~\ref{tab:DE_Member_Selection3} show the checkpoint IDs that reached the highest mIoU. The weights used for the deep ensemble and the weight fusion differ accordingly.
For the determination of the fusion parameter alpha for WF, a grid search in 0.05 steps was performed for each combination.   
Please note that there is no particular reason to select the train instead of the validation data for this selection process. 

 \begin{table}[t]
  \centering
  \caption{Specification of the mIoU (\%) for the selection process of the weights for DE\_2 and WF\_2. The checkpoint IDs (1 to 4) that resulted in the highest mIoU on the training data (marked in bold) were used for the evaluation of the test data.}
  \label{tab:DE_Member_Selection2}
  \begin{tabular}{rcc||cc}
    \toprule
    & \multicolumn{2}{c}{DE\_2}  & \multicolumn{2}{c}{WF\_2}  \\
    \cmidrule(r){2-3}
    \cmidrule(r){4-5}
    \multicolumn{1}{r|}{Checkpoint IDs} & Train data & Test data & Train data & Test data \\
    \midrule
  \multicolumn{1}{r|}{1, 2} & 84,12 & 62.11 & 86.48 & 61.93 \\
  \multicolumn{1}{r|}{1, 3} & 84.06 & 62.25 & 86.62 & 62.38 \\
  \multicolumn{1}{r|}{1, 4} & 83.61 & 62.45 & 86.27 & 62.31 \\
  \multicolumn{1}{r|}{2, 3} & \textbf{84.18} & \textbf{62.04} & 86.64 & 62.18 \\
  \multicolumn{1}{r|}{2, 4} & 83.70 & 62.13 & 86.50 & 61.89 \\
  \multicolumn{1}{r|}{3, 4} & 83.65 & 62.44 & \textbf{86.68} & \textbf{62.17} \\
    \bottomrule
  \end{tabular}
\end{table}

 \begin{table}[t]
  \centering
  \caption{Specification of the mIoU (\%) for the selection process of the weights for DE\_3 and WF\_3. The checkpoint IDs (1 to 4) that resulted in the highest mIoU on the training data (marked in bold) were used for the evaluation of the test data.}
  \label{tab:DE_Member_Selection3}
  \begin{tabular}{rcc||cc}
    \toprule
    & \multicolumn{2}{c}{DE\_3}  & \multicolumn{2}{c}{WF\_3}  \\
    \cmidrule(r){2-3}
    \cmidrule(r){4-5}
    \multicolumn{1}{r|}{Checkpoint IDs} & Train data & Test data & Train data & Test data \\
    \midrule
  \multicolumn{1}{r|}{1, 2, 3} & \textbf{84.29} & \textbf{62.39} & 86.71 & 62.36 \\
  \multicolumn{1}{r|}{1, 2, 4} & 83.99 & 62.52 & 86.57 & 61.96 \\
  \multicolumn{1}{r|}{1, 3, 4} & 83.94 & 62.65 & 86.74 & 62.25 \\
  \multicolumn{1}{r|}{2, 3, 4} & 83.99 & 62.51 & \textbf{86.74} & \textbf{62.27} \\
    \bottomrule
  \end{tabular}
\end{table}

\section{Expected Calibration Error (ECE)}
\label{calibration_ece}

Evaluation of DNN calibration is done using both expected calibration error (ECE) and maximum calibration error (MCE) \cite{guo2017calibration, neumann2018relaxed}. 

To evaluate the calibration metrics the segmentation predictions should be first sorted into fixed-sized bins $\beta_i$, $i =$ 1, . . . , $B$ according to their confidence. Afterwards, for each bin $\beta_i$, accuracy (acc) and confidence (conf) are computed as:

\begin{equation}
\text{acc}_i = \frac{\text{TP}_i}{|\beta _i |}, \qquad \text{conf}_i = \frac{1}{|\beta _i |} \sum_{j=1}^{|\beta _i |} \hat{c}_i
\end{equation}

where $|\beta _i |$ indicates the number of examples in $\beta_i$ and $\hat{c}_i$ is the respective confidence, i.e., softmax score. $\text{TP}_i$ denotes the number of correctly classified in $\beta_i$. The ECE and MCE are then calculated accordingly:

\begin{equation}
    ECE = \frac{1}{B}\sum_{i=1}^{B}|\text{acc}_i - \text{conf}_i|,    \qquad  MCE = max |\text{acc}_i - \text{conf}_i|.
\end{equation}

\section{Precision, Recall, KL}
\label{precision_recall_kl}

\begin{figure}[t]
    \centering
    \subfigure[]{\includegraphics[width=0.45\textwidth]{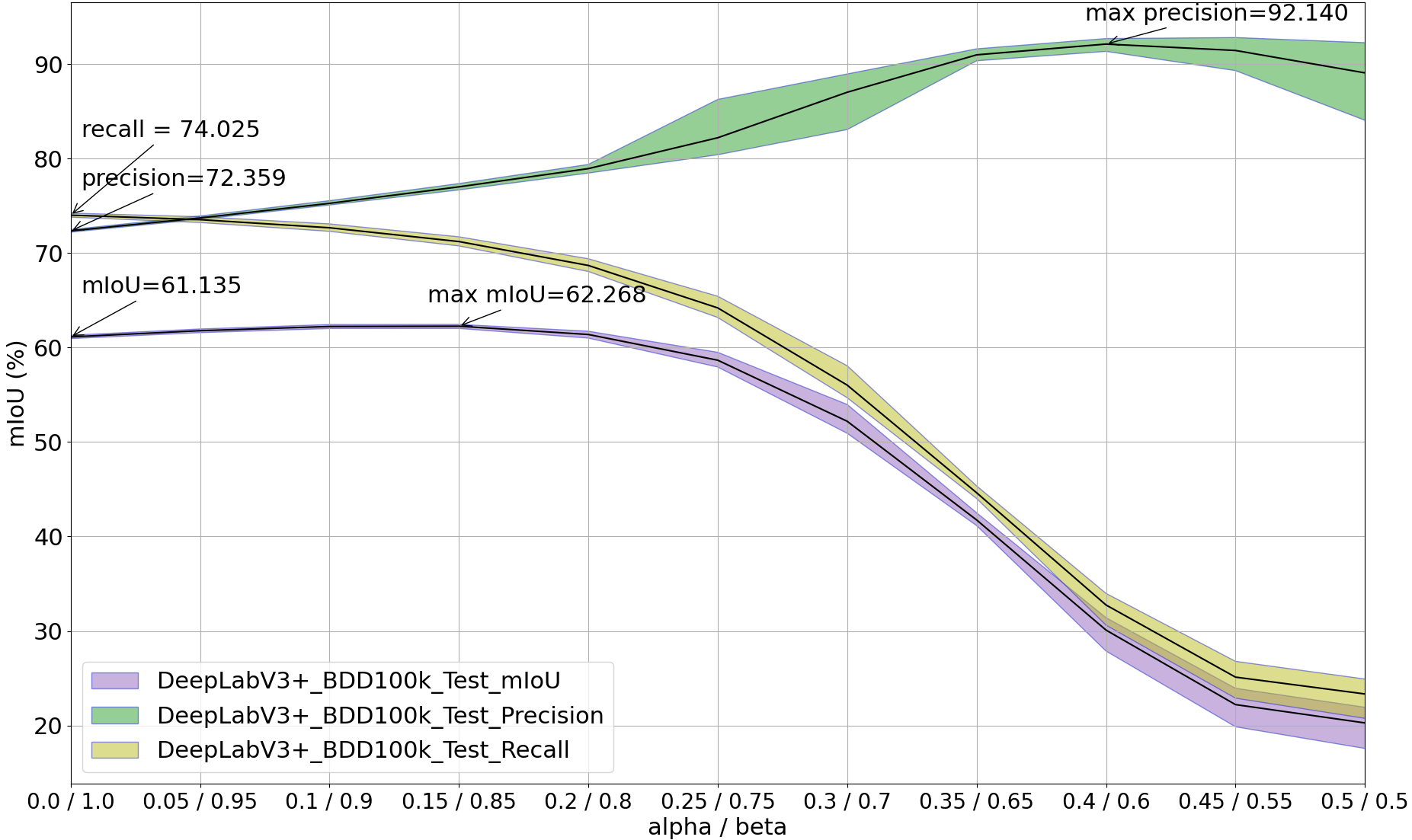}}
    \quad
    \subfigure[]{\includegraphics[width=0.45\textwidth]{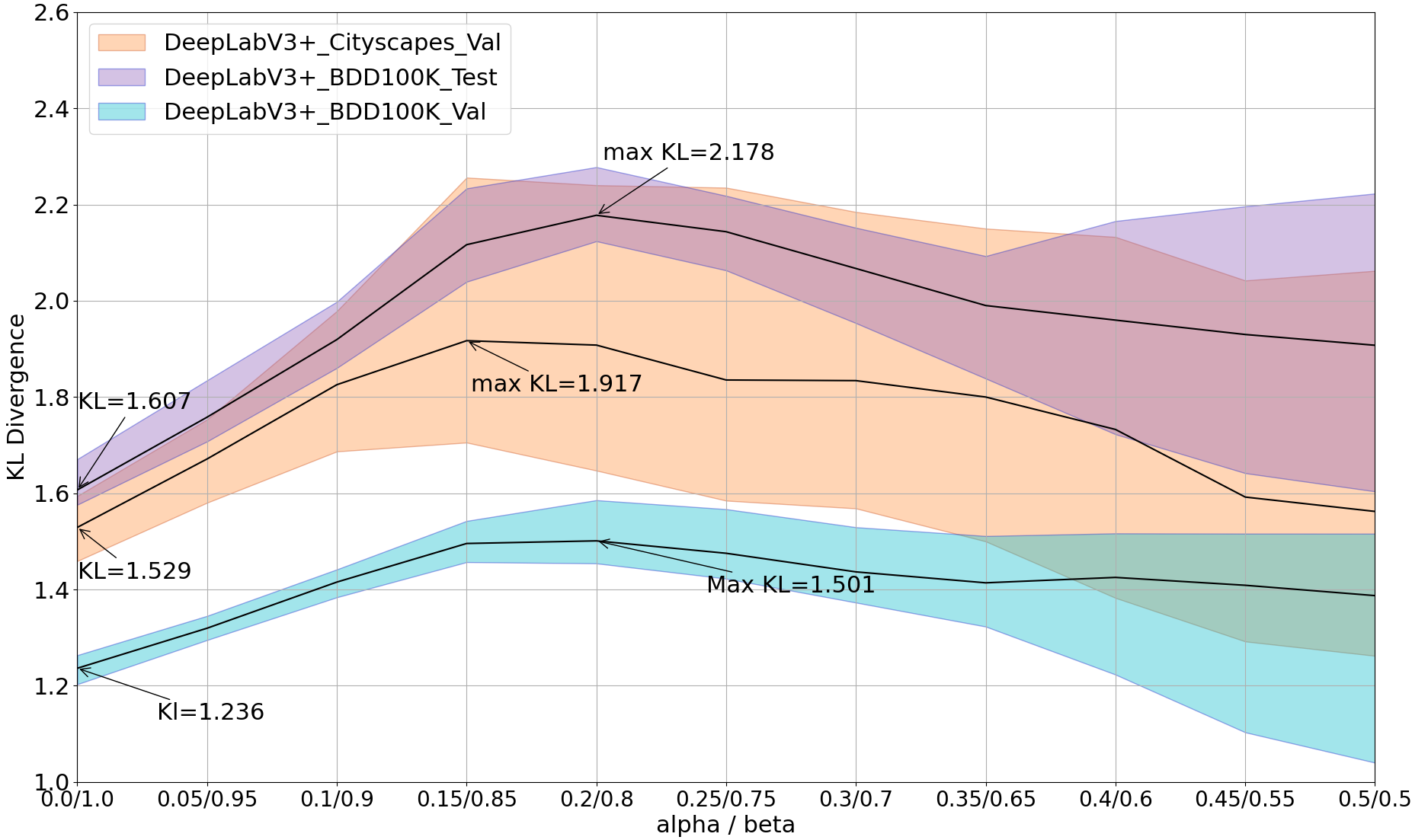}}
    \caption{DeepLabV3+ performance (precision, recall) and calibration (KL) for 6 fused weights averaged for alpha from 0 to 1 in 0.05 steps. Please note that alpha/beta = 0/1 corresponds to the single weights. \textit{(a)} precision and recall \textit{(b)} KL.}
    \label{fig:performance_from_sratch_2}
\end{figure}

Fig.~\ref{fig:performance_from_sratch_2}(a) shows the precision and recall for DeepLabV3+ on the BDD100K data. Interestingly, precision increases significantly along the x-axis, while recall decreases sharply. 

Furthermore, the ability to separate between correct and incorrect classifications is an important factor in our tests. For that, we look into calculating the network’s ability in being able to differentiate between correct and incorrect labels (distinctive distributions) in~\ref{fig:performance_from_sratch_2}(b). Accordingly, the distributions of both correct and incorrect predictions are plotted with respect to confidence score, then both distributions are compared using the Kullback-Leibler (KL) divergence metric~\cite{kullback1951information} as follows:

\begin{equation}
D_{KL}(P \parallel Q) = - \sum_{x \in \mathcal{X}} P(x) \text{log} \left(\frac{P(x)}{Q(x)} \right), 
\end{equation}
where $P$ and $Q$ represent the correct and incorrect distributions respectively, and $\mathcal{X}$ represents the confidence score space.

High KL value indicates dissimilar distinctive distributions reflecting the efficiency of the DNN and vice versa for low KL divergence value, whilst the value 0 indicates the inability of using the confidence score in distinguishing between correct and incorrect predictions. Figure~\ref{fig:performance_from_sratch_2}(b) shows that the fusion also improves the KL divergence achieving higher value than the initial weights.

\section{Additional Quantitative and Qualitative Results for WF and SWA Comparison}\label{more_visual_results}

Table~\ref{tab:Comparison_to_SWA2} shows the per class IoU for the comparison between the single weight, SWA and WF. On the BDD100K validation data, SWA is 0.77~\% mIoU better than the single weight. WF exceeds this with an improvement of 1.44~\% mIoU. On the test data, the picture is similar. Again, WF outperforms the single weight and SWA. It should be noted again that for WF the fusion parameter alpha has been determined on the validation data using grid search in 0.05 steps. It shows that the increased mIoU by WF generalizes to the test data as well. 
Looking at the per class IoU it is noticeable that mainly classes with a smaller pixel density such as traffic light, traffic sign, pole etc. benefit the most from the fusion. This applies to WF and SWA. 

\begin{figure}[t]
    \includegraphics[width=1\textwidth]{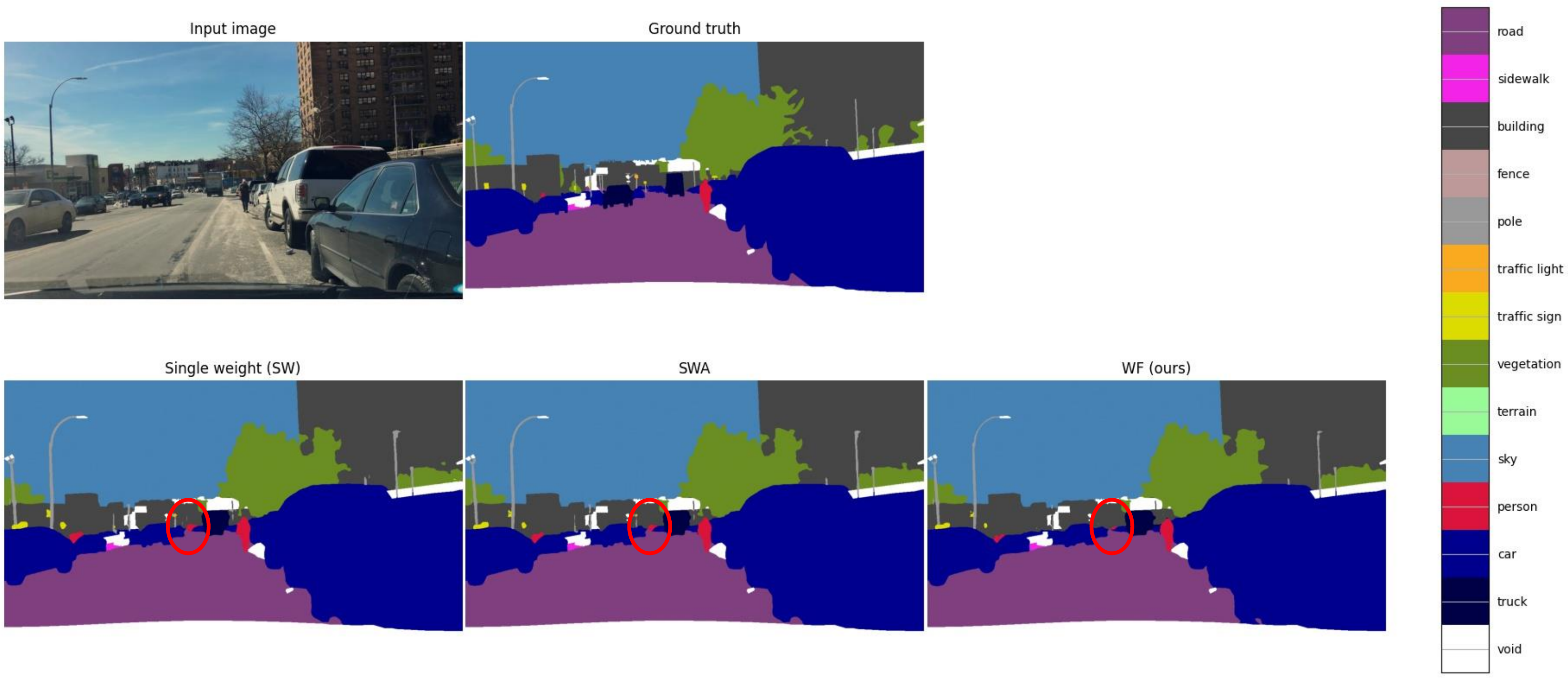}
	\centering
	\caption{Comparison of semantic segmentation masks between SW, SWA and WF. The areas to be pointed out are marked with a red ellipse.}
	\label{fig:swa2_cropped}
\end{figure}

\begin{figure}[h!]
    \includegraphics[width=1\textwidth]{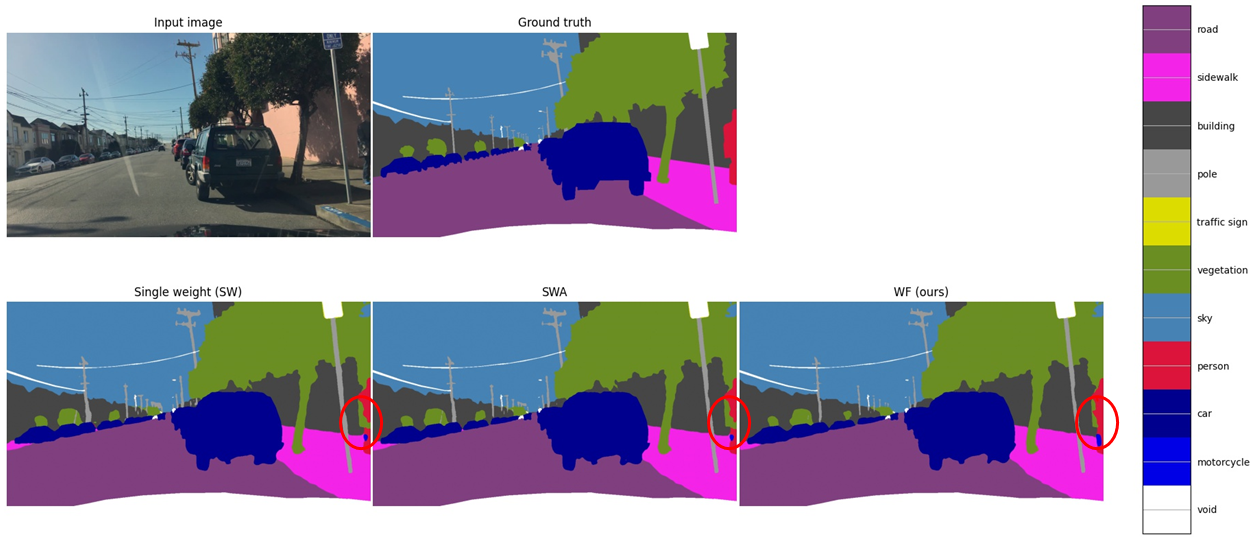}
	\centering
	\caption{Comparison of semantic segmentation masks between SW, SWA and WF. The areas to be pointed out are marked with a red ellipse.}
	\label{fig:swa3_cropped}
\end{figure}

Figure~\ref{fig:swa2_cropped} and \ref{fig:swa3_cropped} show further visual comparisons of SW, SWA, and WF. The areas to be pointed out are marked with a red ellipse. In figure~\ref{fig:swa2_cropped} we see false positive segmentation from SW and SWA for the \glqq person\grqq{} class. In figure~\ref{fig:swa3_cropped}, on the other hand, the \glqq person\grqq{} class is segmented more completely by WF. 

\begin{table}[t]

\caption{Per class mIoU in (\%) for DeepLabV3+ and the BDD100K data.}
\scriptsize
 \centering
 \label{tab:Comparison_to_SWA2}
 \resizebox{\textwidth}{!}{%
\begin{tabular}{cccc|ccc}
\toprule
                                   & \multicolumn{3}{c}{BDD100K validation data}                                        & \multicolumn{3}{c}{BDD100K test data}               \\ \hline
\multicolumn{1}{r|}{Classes}       & Single weight           & SWA    & \multicolumn{1}{c|}{WF (ours)}           & Single weight           & SWA   & WF (ours)           \\ \hline

\multicolumn{1}{r|}{Road} & 94.48 & 94.53\color{ForestGreen}(+0.05) & 94.79\color{ForestGreen}(+0.26) & 94.53 & 94.53\color{ForestGreen}(0.00) & 94.67\color{ForestGreen}(+0.14) \\       
\multicolumn{1}{r|}{Sidewalk} & 63.48 & 63.25\color{red}(-0.23) & 64.66\color{ForestGreen}(+1.18) & 64.57 & 64.66\color{ForestGreen}(+0.09) & 65.59\color{ForestGreen}(+1.02)  \\   
\multicolumn{1}{r|}{Building} & 83.08 & 83.29\color{ForestGreen}(+0.21) & 83.40\color{ForestGreen}(+0.32) & 85.42 & 85.72\color{ForestGreen}(+0.3) & 85.69\color{ForestGreen}(+0.27) \\   
\multicolumn{1}{r|}{Wall} & 37.50 & 42.66\color{ForestGreen}(+5.16) & 43.33\color{ForestGreen}(+5.83) & 26.99 & 31.00\color{ForestGreen}(+4.01) & 29.13\color{ForestGreen}(+2.14) \\    
\multicolumn{1}{r|}{Fence} & 46.85 & 46.92\color{ForestGreen}(+0.07) & 46.58\color{red}(-0.27) & 50.24 &  50.95\color{ForestGreen}(+0.71) & 51.21\color{ForestGreen}(+0.97) \\    
\multicolumn{1}{r|}{Pole} & 47.00 & 50.00\color{ForestGreen}(+3.00) & 50.25\color{ForestGreen}(+3.25) & 49.71 & 52.68\color{ForestGreen}(+2.97) & 52.91\color{ForestGreen}(+3.20)  \\   
\multicolumn{1}{r|}{Traffic Light} & 49.02 & 53.42\color{ForestGreen}(+4.40) & 53.55\color{ForestGreen}(+4.53) & 53.04 & 57.16\color{ForestGreen}(+4.12) & 58.14\color{ForestGreen}(+5.10) \\ 
\multicolumn{1}{r|}{Traffic Sign} & 58.78 & 61.54\color{ForestGreen}(+2.76) & 62.66\color{ForestGreen}(+3.88) & 53.16 & 55.65\color{ForestGreen}(+2.49) & 56.39\color{ForestGreen}(+3.23) \\ 
\multicolumn{1}{r|}{Vegetation} & 85.11 & 85.48\color{ForestGreen}(+0.37) & 85.44\color{ForestGreen}(+0.33) & 86.32 & 86.29\color{red}(-0.03) & 86.43\color{ForestGreen}(+0.11) \\ 
\multicolumn{1}{r|}{Terrain} & 45.76 & 46.94 \color{ForestGreen}(+1.18) & 45.87\color{ForestGreen}(+0.11) & 50.42 & 50.98\color{ForestGreen}(+0.56) & 51.00\color{ForestGreen}(+0.58) \\ 
\multicolumn{1}{r|}{Sky} & 95.64 & 95.72\color{ForestGreen}(+0.08) & 95.62\color{red}(-0.02) & 95.20 & 95.09\color{red}(-0.11) & 95.20\color{ForestGreen}(0.00)  \\     
\multicolumn{1}{r|}{Person} & 62.70 & 63.42\color{ForestGreen}(+0.72) & 65.25\color{ForestGreen}(+2.55) & 63.80 & 64.67\color{ForestGreen}(+0.87) & 65.14\color{ForestGreen}(+1.34)   \\   
\multicolumn{1}{r|}{Rider} & 45.93 & 42.22\color{red}(-3.71) & 48.09\color{ForestGreen}(+2.16) & 50.07 & 51.06\color{ForestGreen}(+0.99) & 52.48\color{ForestGreen}(+2.41)  \\   
\multicolumn{1}{r|}{Car} & 88.10 & 88.63\color{ForestGreen}(+0.53) & 88.73\color{ForestGreen}(+0.63) & 90.37 & 90.44\color{ForestGreen}(+0.07) & 90.79\color{ForestGreen}(+0.42)  \\       
\multicolumn{1}{r|}{Truck} & 59.18 & 62.35\color{ForestGreen}(+3.17) & 61.56\color{ForestGreen}(+2.38) & 57.49 & 59.57\color{ForestGreen}(+2.08) & 60.55\color{ForestGreen}(+3.06) \\      
\multicolumn{1}{r|}{Bus} & 73.36 & 74.66\color{ForestGreen}(+1.30) & 70.47\color{red}(-2.89) & 78.93 & 78.90\color{red}(-0.03) & 80.66\color{ForestGreen}(+1.73) \\     
\multicolumn{1}{r|}{Train} & 0 & 0 & 0 &  0 & 0 & 0 \\    
\multicolumn{1}{r|}{Motorcycle} & 35.09 & 32.14\color{red}(-2.95) & 36.25\color{ForestGreen}(+1.16) & 55.46 & 52.70\color{red}(-2.76) & 54.24\color{red}(-1.22) \\ 
\multicolumn{1}{r|}{Bicycle} & 42.60 & 41.78\color{red}(-0.82) & 44.52\color{ForestGreen}(+1.92) & 49.75 & 48.68\color{red}(-1.07) & 48.78\color{red}(-0.97)  \\  \hline
\multicolumn{1}{r|}{Overall} & 58.61 & 59.38\color{ForestGreen}(+0.77) & 60.05\color{ForestGreen}(+1.44) &  60.81 & 61.67\color{ForestGreen}(+0.86) & 62.05\color{ForestGreen}(+1.24) 
\end{tabular}}
\end{table}

\end{document}